\documentclass{article}

\usepackage{iclr2025_conference,times}
\iclrfinalcopy

\renewcommand{\headrulewidth}{0pt}
\usepackage{amsmath,amsfonts,bm}

\def\eqref#1{equation~\ref{#1}}

\def\1{\bm{1}}

\DeclareMathAlphabet{\mathsfit}{\encodingdefault}{\sfdefault}{m}{sl}
\SetMathAlphabet{\mathsfit}{bold}{\encodingdefault}{\sfdefault}{bx}{n}

\usepackage{hyperref}
\usepackage{url}

\usepackage{graphicx}
\usepackage{booktabs}
\usepackage{amsmath}
\usepackage{amssymb}
\usepackage{subcaption}
\usepackage{multirow}
\usepackage{array}
\usepackage{booktabs}
\usepackage{tikz}

\title{
Olfactory-Inspired Sparse Combinatorial Coding for Low-Resource Named Entity Recognition\thanks{Code available at \url{https://github.com/bhushan1729/olfaction-inspired-ner}}
}

\author{
Bhushan Deshpande \\
Independent Researcher, India \\
\texttt{bhushan1729dsp@gmail.com}
}

\begin{document}

\maketitle
\lhead{}

\begin{abstract}
Named Entity Recognition (NER) in low-resource languages suffers from limited supervision and a lack of high-quality pretrained embeddings. Biological olfaction, which relies on sparse combinatorial coding through receptor and glomerular organization, offers a compelling paradigm for learning robust representations under uncertainty. In this paper, we introduce a receptor-glomerular bottleneck, a novel, biologically-inspired olfactory architecture, between standard token embeddings and a BiLSTM-CRF sequence model. We evaluate our architecture across six multilingual datasets trained entirely from scratch (without pre-trained embeddings) under varied data-scale conditions, including a strict 1k-sentence low-resource control. Our results demonstrate that introducing a representation bottleneck yields F1 score improvements under severe data scarcity, primarily by acting as a powerful regularizer. Under the 1k capped training condition, at least one olfactory-inspired configuration achieves the highest mean F1 score across all six datasets. While these improvements represent near-ties with generic bottleneck controls for most languages, the olfactory architecture provides a significant advantage in languages like Bangla (+6.23\% F1 over standard baseline and +8.47\% F1 over the best control baseline) where generic bottlenecks degrade performance. We also observe improvements in the ultra-low-resource Telugu setting (+4.43\% F1) at full-scale, and find that sparse specialization naturally emerges within the receptor layer. Our findings suggest that structured sparse coding inspired by olfactory networks serves as an effective inductive bias and regularizer when representations must be learned from limited or noisy supervision.
\end{abstract}

\section{Introduction}

Named Entity Recognition (NER) in low-resource languages remains challenging due to limited annotated data and the difficulty of learning robust representations from scarce supervision \citep{sunna2023languagefamilies,jia2021fewshot}. Although transformer-based models achieve strong performance in resource-rich settings \citep{devlin2019bert}, their effectiveness often diminishes for underrepresented and morphologically complex languages where training data is limited.

Biological olfactory systems employ a distributed coding strategy in which individual receptors respond to multiple odorants and individual odorants activate overlapping receptor populations. These receptor responses converge onto glomeruli, producing sparse combinatorial activity patterns that enable robust odor discrimination and efficient representation of high-dimensional sensory inputs \citep{buck1991novel,lin2014sparse,wang2021evolving}.

We hypothesize that sparse combinatorial coding may provide a useful inductive bias for low-resource NER. To test this, we introduce an exploratory architecture that inserts a receptor-glomerular bottleneck into a standard BiLSTM-CRF model. 

Our contributions are as follows:
\begin{enumerate}
    \item We introduce a receptor--glomerular bottleneck architecture for NER.
    \item We evaluate this architecture across six multilingual datasets with varying resource levels.
    \item We show that when training from scratch under low-resource constraints (1k sentences), at least one olfactory-inspired configuration achieves the highest mean F1 score across all six languages, with the largest improvements occurring in the lowest-resource settings (Bangla and English) while stabilizing training variance across random seeds.
    \item We demonstrate the natural emergence of sparse receptor specialization.
    \item We analyze the conditions under which this inductive bias helps and fails.
\end{enumerate}

\section{Related Work}

\subsection{Named Entity Recognition}
NER is traditionally modeled as sequence labeling. Standard architectures employ BiLSTM-CRF \cite{huang2015bilstmcrf} and transformer-based methods \cite{devlin2019bert, lample2016neural}. Multilingual and low-resource NER typically focuses on cross-lingual transfer, data augmentation, or prompt engineering, as high-capacity models rely heavily on large-scale supervision. A comprehensive survey of modern low-resource and multilingual NER paradigms is provided by \cite{keraghel2024recent}, classifying approaches into cross-lingual transfer, data-efficient architecture optimization, and LLM prompting. Recent low-resource NER work has also explored cross-lingual transfer from related language families \cite{sunna2023languagefamilies} and meta-learning for few-shot scenarios \cite{jia2021fewshot}, emphasizing that structural inductive bias is a key component to fast generalization from limited examples.

\subsection{Sparse Representations}
Sparse coding, mixture-of-experts (MoE), and structured bottlenecks are widely used for feature disentanglement, capacity control, and general representation learning \cite{goodfellow2016deep}. Classic neuroscience research has shown that enforcing sparsity yields interpretable codes, such as sparse coding of natural images producing edge detectors \cite{olshausen1996sparse}. In deep learning, Shazeer et al. \cite{shazeer2017moe} introduced a sparsely-gated MoE layer that routes inputs to a few expert sub-networks, echoing our use of receptors as fixed sparse experts. More recently, sparse mixture models like the Switch Transformer \cite{fedus2022switch} have demonstrated that routing inputs to sparse, specialized sub-networks increases model capacity while keeping computational cost constant. Furthermore, Yang et al. \cite{yang2025structuredib} proposed a Structured Information Bottleneck to preserve relevant information under compression. Our work shares similarities with these methods by enforcing sparse activation, but it specifically targets a combinatorial feature aggregation step loosely inspired by olfactory wiring.

\subsection{Neuroscience-Inspired AI}
AI has frequently drawn from neuroscience, including attention mechanisms (cognition) \citep{vaswani2017attention}, predictive coding \citep{rao1999predictive}, and hippocampal memory systems \citep{graves2016hybrid}. Previous olfactory computation literature has explored robustness and associative learning \citep{babadi2014sparseness,aso2014neuronal}. A striking example of artificial systems converging on biological structures is the work by Wang et al. \citep{wang2021evolving}, who demonstrated that a neural network trained on an odor classification task spontaneously developed a receptor-glomeruli architecture mirroring biological olfaction. This supports the notion that sparse, combinatorial layers can emerge naturally under pressure to compress features. We emphasize that our architecture is an abstract computational analogy to olfactory processing, not a biological simulation.

\subsection{Information Bottleneck and Representation Compression}
Our proposed sparse bottleneck is closely linked to the Information Bottleneck (IB) framework \cite{tishby2000ib}. The IB method formalizes the trade-off between compressing the input representation $Z$ of a source variable $X$ and preserving its predictive capacity with respect to a target variable $Y$. Variational approximations such as the Deep Variational Information Bottleneck (VIB) \cite{alemi2017vib} employ variational inference to parameterize this bottleneck in deep neural networks. More recently, Yang et al. \cite{yang2025structuredib} introduced the Structured Information Bottleneck to enforce structural priors on compressed representations. By introducing an explicit, sparse receptor-glomerular projection layer, our model acts as a structured bottleneck. It forces the network to compress high-dimensional word representations into a lower-dimensional, non-negative sparse activation space, discarding task-irrelevant, volatile features while retaining essential semantic signals.

\section{Biological Motivation}

\subsection{The Olfactory Pathway}
In biological olfactory systems (found in both vertebrates and insects), odor detection and processing follow a highly structured, conserved pathway that maps chemical stimuli to neural representations:
\begin{enumerate}
    \item \textbf{Sensory Neurons (OSNs/ORNs):} Olfactory sensory neurons (OSNs) express exactly one type of olfactory receptor from a large multigene family \citep{vosshall1999spatial, vosshall2000olfactory}. Each receptor responds selectively to specific chemical features (epitopes) of odor molecules. In mice, for example, each sensory neuron expresses only one of $\sim 1,000$ odorant receptors \citep{buck1991novel, godfrey2004mouse, zhang2002olfactory}.
    \item \textbf{Glomerular Convergence:} All OSNs expressing the same specific receptor converge onto an anatomically distinct locus called a glomerulus (located in the antennal lobe of insects, or the olfactory bulb of vertebrates) \citep{vosshall1999spatial, vosshall2000olfactory, mombaerts1996visualizing, ressler1993zonal, ressler1994information, vassar1994topographic}. This acts as a severe structural bottleneck, pooling many redundant inputs to filter noise and amplify signals.
    \item \textbf{Projection \& Sharpening (Mitral Cells / Projection Neurons):} Glomerular activations are processed and relayed by principal output neurons---mitral/tufted cells in vertebrates, or projection neurons (PNs) in insects. Most PNs innervate a single glomerulus \citep{jefferis2007comprehensive, marin2002representation, wong2002spatial}, while mitral/tufted cells project to the primary olfactory cortex \citep{price1970mitral}. These cells can refine and sharpen combinatorial activation patterns (often through lateral inhibition).
    \item \textbf{Higher Cortical Processing:} These output neurons project to higher brain regions---such as the piriform cortex in mammals (where they synapse onto $\sim 1$ million piriform neurons) or the mushroom body (MB)/Kenyon cells (KCs) in insects---where sparse combinatorial codes are translated into associative memories, patterns, and behavioral decisions \citep{deBelle1994associative, dubnau2001disruption, heisenberg1985drosophila, mcguire2001role, davison2011neural, miyamichi2011cortical}. Individual KCs receive unstructured input from $\sim 4-10$ PNs \citep{caron2013random, li2020connectome, zheng2018complete}, and piriform neurons receive roughly $30-100$ inputs from a random collection of glomeruli \citep{davison2011neural, miyamichi2011cortical}.
\end{enumerate}

\subsection{Key Computational Properties}
\begin{itemize}
    \item \textbf{Sparse Activation:} Only a small subset of olfactory receptors fires for any given odorant, leading to highly efficient energy and representation usage \citep{caron2013random, li2020connectome}.
    \item \textbf{Combinatorial Coding:} Meaning is encoded combinatorially; individual receptors are broad/weak feature detectors, and the identity of an odor is determined by the specific combination of activated receptors rather than a single ``labeled line'' \citep{wang2021evolving}.
    \item \textbf{Robustness and Noise Tolerance:} The convergent pooling of thousands of sensory neurons into a small number of glomeruli averages out stochastic noise, allowing the system to detect weak signals in complex backgrounds \citep{ressler1993zonal, ressler1994information, vassar1994topographic}.
    \item \textbf{Emergent Specialization:} Different receptors develop sensitivity to distinct molecular features, establishing a distributed feature extraction system \citep{buck1991novel, godfrey2004mouse, zhang2002olfactory}.
\end{itemize}

\subsection{Mapping to NLP}
This biological architecture provides an intuitive blueprint for sequence labeling tasks like Named Entity Recognition. Table~\ref{tab:olfactory_mapping} summarizes the mapping between biological olfactory components and the proposed NER architecture. This mapping is an abstract computational analogy inspired by biological olfactory wiring \citep{wang2021evolving} rather than a direct physiological simulation.

\begin{table}[ht]
\centering
\small
\resizebox{\columnwidth}{!}{
\begin{tabular}{lll}
\toprule
\textbf{Biological Olfactory System} & \textbf{NLP/NER Equivalent} & \textbf{Function} \\
\midrule
Odor molecules / Chemical stimuli & Token embeddings & Raw sensory input \\
Olfactory Sensory Neurons (OSNs) & Receptor Layer & Sparse, localized feature detection \\
Glomeruli (Olfactory Bulb / Antennal Lobe) & Glomerular Layer & Convergent feature pooling and noise reduction \\
Mitral Cells / Projection Neurons & Mitral Layer (optional) & Output projection and feature sharpening \\
Olfactory Cortex / Mushroom Body & BiLSTM Encoder & Contextual sequence encoding \\
Higher Behavioral Output & CRF Decoder & Sequence decoding and tag assignment \\
\bottomrule
\end{tabular}
}
\caption{Mapping between the biological olfactory system and the proposed NER architecture.}
\label{tab:olfactory_mapping}
\end{table}

\section{Methodology}

\subsection{Baseline Architecture}
Our baseline is a standard sequence tagger: Embedding $\rightarrow$ BiLSTM $\rightarrow$ CRF. The word embeddings have a dimensionality of $d=300$. These are fed into a 1-layer bidirectional LSTM with a hidden dimension of $256$. The outputs are projected to the target label space and decoded using a Conditional Random Field (CRF) layer. Total parameter count for the baseline model is approximately 1.5 million (excluding the embedding matrix).

\subsection{Olfactory Architecture}
The olfactory-enhanced architecture introduces a biologically-inspired sparse bottleneck between the embeddings and the BiLSTM. The forward pass is defined as: Embedding $\rightarrow$ Receptor Layer $\rightarrow$ Glomerular Layer $\rightarrow$ BiLSTM $\rightarrow$ CRF.

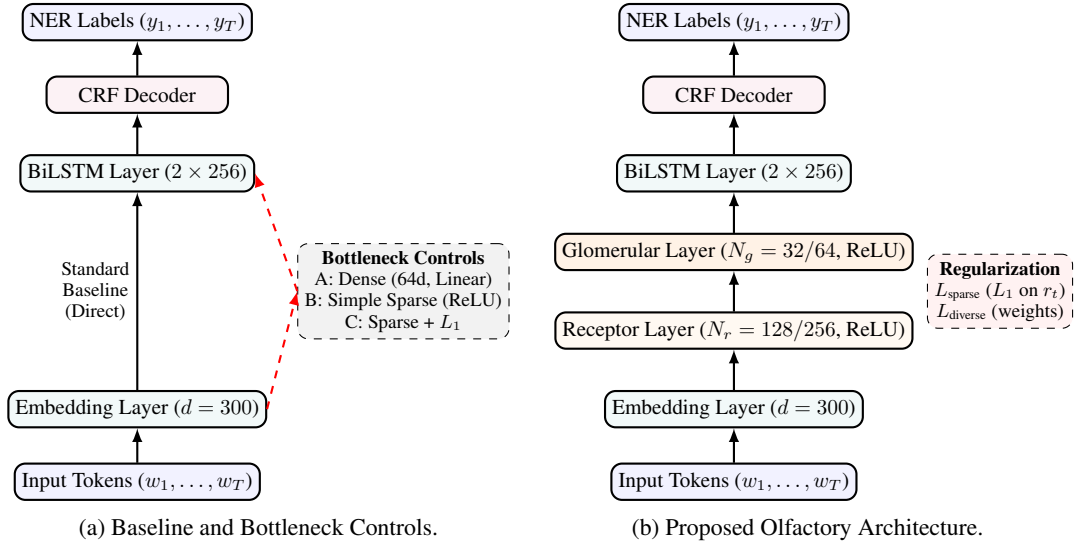
\begin{figure*}[htbp]
\centering
\begin{subfigure}[b]{0.48\textwidth}
\centering
\begin{tikzpicture}[
    box/.style={draw, rectangle, minimum width=3.0cm, minimum height=0.6cm, align=center, rounded corners, fill=blue!5, thick},
    arrow/.style={-latex, thick},
    scale=0.8, every node/.style={scale=0.8}
]
\node (input) at (0,0) [box] {Input Tokens ($w_1, \dots, w_T$)};
\node (embedding) at (0,1.2) [box, fill=teal!5] {Embedding Layer ($d=300$)};
\node (bilstm) at (0,5.1) [box, fill=teal!5] {BiLSTM Layer ($2 \times 256$)};
\node (crf) at (0,6.4) [box, fill=purple!5] {CRF Decoder};
\node (output) at (0,7.6) [box] {NER Labels ($y_1, \dots, y_T$)};

\node (bottleneck) at (4.4, 3.15) [draw, dashed, rectangle, minimum width=2.4cm, minimum height=1.0cm, rounded corners, fill=gray!10, align=center, font=\small] {
    \textbf{Bottleneck Controls} \\
    A: Dense (64d, Linear) \\
    B: Simple Sparse (ReLU) \\
    C: Sparse + $L_1$
};

\draw [arrow] (input) -- (embedding);
\draw [arrow] (embedding) -- node[left, align=center, font=\small] {Standard\\Baseline\\(Direct)} (bilstm);
\draw [arrow, dashed, red] (embedding.east) -- node[below, sloped, font=\tiny] {} (bottleneck.west);
\draw [arrow, dashed, red] (bottleneck.west) -- (bilstm.east);
\draw [arrow] (bilstm) -- (crf);
\draw [arrow] (crf) -- (output);
\end{tikzpicture}
\caption{Baseline and Bottleneck Controls.}
\label{fig:baseline_arch}
\end{subfigure}
\hfill
\begin{subfigure}[b]{0.48\textwidth}
\centering
\begin{tikzpicture}[
    box/.style={draw, rectangle, minimum width=3.0cm, minimum height=0.6cm, align=center, rounded corners, fill=blue!5, thick},
    arrow/.style={-latex, thick},
    scale=0.8, every node/.style={scale=0.8}
]
\node (input) at (0,0) [box] {Input Tokens ($w_1, \dots, w_T$)};
\node (embedding) at (0,1.2) [box, fill=teal!5] {Embedding Layer ($d=300$)};
\node (receptor) at (0,2.5) [box, fill=orange!5] {Receptor Layer ($N_r = 128/256$, ReLU)};
\node (glomeruli) at (0,3.8) [box, fill=orange!10] {Glomerular Layer ($N_g = 32/64$, ReLU)};
\node (bilstm) at (0,5.1) [box, fill=teal!5] {BiLSTM Layer ($2 \times 256$)};
\node (crf) at (0,6.4) [box, fill=purple!5] {CRF Decoder};
\node (output) at (0,7.6) [box] {NER Labels ($y_1, \dots, y_T$)};

\draw [arrow] (input) -- (embedding);
\draw [arrow] (embedding) -- (receptor);
\draw [arrow] (receptor) -- (glomeruli);
\draw [arrow] (glomeruli) -- (bilstm);
\draw [arrow] (bilstm) -- (crf);
\draw [arrow] (crf) -- (output);

\node (losses) at (4.4,3.15) [draw, dashed, rectangle, rounded corners, fill=red!5, align=center, font=\small, minimum width=2.4cm] {
    \textbf{Regularization} \\
    $L_{\text{sparse}}$ ($L_1$ on $r_t$) \\
    $L_{\text{diverse}}$ (weights)
};
\end{tikzpicture}
\caption{Proposed Olfactory Architecture.}
\label{fig:olfactory_arch}
\end{subfigure}
\caption{Architectural flow comparison. (a) Standard BiLSTM-CRF baseline alongside the generic bottleneck controls (A, B, and C). (b) Proposed olfactory-inspired architecture featuring a two-stage sparse bottleneck (Receptor and Glomerular layers) and regularization losses.}
\label{fig:architectures_comparison}
\end{figure*}

\paragraph{Receptor Layer:}
This layer comprises $N_r = 128$ (or $256$) sparse nonlinear projections acting as weak feature detectors. Given an input embedding $x_t \in \mathbb{R}^{300}$, the receptor activation vector $r_t \in \mathbb{R}^{N_r}$ is computed as:
\begin{equation}
r_t = \sigma(W_R x_t + b_R)
\end{equation}
where $W_R \in \mathbb{R}^{N_r \times 300}$ is a dense weight matrix, $b_R \in \mathbb{R}^{N_r}$ is a bias vector, and $\sigma$ is the ReLU activation function \citep{Nair2010RectifiedLU}. The use of ReLU is critical as it naturally enforces a non-negative, sparse firing pattern akin to biological olfactory receptors.

\paragraph{Glomerular Layer:}
Receptors aggregate their signals into a smaller number of glomeruli ($N_g = 32$ or $64$), acting as convergent feature pooling and noise reduction. The glomerular activation vector $g_t \in \mathbb{R}^{N_g}$ is computed as:
\begin{equation}
g_t = \text{ReLU}(W_G r_t)
\end{equation}
where $W_G \in \mathbb{R}^{N_g \times N_r}$ serves as the assignment matrix defining the connection strength from receptors to glomeruli. The output $g_t$ is then passed into the BiLSTM (hidden dimension $256$).

\paragraph{Sparsity and Diversity Regularization:}
To encourage distinct, specialized receptor functions and prevent redundant feature collapse, we optimize the network using a composite loss function:
\begin{equation}
L = L_{\text{NER}} + \lambda_{\text{sparse}} L_{\text{sparse}} + \lambda_{\text{diverse}} L_{\text{diverse}}
\end{equation}
Here, $L_{\text{NER}}$ is the standard negative log-likelihood from the CRF. 

The sparsity loss $L_{\text{sparse}}$ acts as an $L_1$ penalty on the receptor activations to enforce population sparsity, computed over a sequence of length $T$ as:
\begin{equation}
L_{\text{sparse}} = \frac{1}{T} \sum_{t=1}^T \|r_t\|_1
\end{equation}
where $r_t \in \mathbb{R}^{N_r}$ is the receptor activation vector at time step $t$.

The diversity loss $L_{\text{diverse}}$ penalizes pairwise cosine similarity between the weight vectors of different receptors to prevent redundant representation collapse, computed as:
\begin{equation}
L_{\text{diverse}} = \frac{1}{N_r(N_r-1)} \sum_{i \neq j} \frac{|W_{R,i} \cdot W_{R,j}|}{\|W_{R,i}\|_2 \|W_{R,j}\|_2}
\end{equation}
where $W_{R,i}$ denotes the weight vector (the $i$-th row of $W_R$) for the $i$-th receptor. We set the regularization coefficients to $\lambda_{\text{sparse}} = 0.001$ and $\lambda_{\text{diverse}} \in \{0.01, 0.05\}$.

\paragraph{Training Procedure:}
Models are trained using the Adam optimizer \citep{Kingma2015AdamAM} with a learning rate of $0.001$. We use a batch size of $32$ and train for up to $30$ epochs, applying early stopping with a patience of $5$ epochs based on validation F1-score. Dropout ($p=0.2$) is applied after the embedding layer and before the BiLSTM.

\paragraph{Relationship to Control Baselines:}
Our formulation naturally generalizes the control baselines by setting specific constraints on the layers and loss terms. In particular, Baseline A (Dense Bottleneck) is recovered by removing the Glomerular layer, setting the receptor count $N_r = 64$, using an identity activation function ($\sigma = \text{Identity}$), and disabling regularization ($\lambda_{\text{sparse}} = 0$, $\lambda_{\text{diverse}} = 0$). Baseline B (Simple Sparse Bottleneck) is obtained similarly but retaining the ReLU activation ($\sigma = \text{ReLU}$). Baseline C (Sparse Bottleneck with $L_1$) further adds the activation penalty by setting $\lambda_{\text{sparse}} = 0.001$. This uniform representation allows us to isolate the effects of two-stage mapping, non-negativity, and regularization within a single unified framework.

\subsection{Receptor Selectivity Index (RSI)}
Existing metrics in representation learning typically quantify overall sparsity (e.g., the proportion of active units) or variance, but they fail to measure whether individual units develop functional specialization to semantic categories. To address this limitation and directly quantify the interpretability of our learned sparse representations, we introduce the Receptor Selectivity Index (RSI). RSI measures the degree to which a specific unit (e.g., an individual receptor or glomerulus) is specialized to detect particular named entity classes rather than firing uniformly across all classes.

For a given unit $r$, let $\mu_{r, e}$ represent the mean activation of that unit when exposed to tokens belonging to entity type $e \in \mathcal{E}$ (e.g., PER, LOC, ORG). The RSI is formulated as the normalized range of its mean activations across all entity types:
\begin{equation}
RSI(r) = \begin{cases} 
\frac{\max_{e}(\mu_{r, e}) - \min_{e}(\mu_{r, e})}{\max_{e}(\mu_{r, e})} & \text{if } \max_{e}(\mu_{r, e}) > 10^{-6} \\ 
0 & \text{otherwise} 
\end{cases}
\end{equation}
where $10^{-6}$ is a small threshold to avoid division by zero for inactive units. 

An RSI near $1.0$ indicates extreme specialization (the unit fires strongly for at least one entity type and is nearly silent for at least one other), while an RSI near $0.0$ implies a lack of selectivity (the unit fires uniformly regardless of the entity class).

\section{Experimental Setup}

\subsection{Datasets}
We evaluate our proposed model on the standard CoNLL-2003 English dataset \cite{tjongkimsang2003conll} and five low-resource/multilingual language datasets from WikiANN (Marathi, Hindi, Tamil, Bangla, and Telugu). Table~\ref{tab:datasets} summarizes the dataset sizes, resource levels, and embedding initialization schemes. Crucially, all models are evaluated under strict representation learning limits where embeddings are trained randomly from scratch.

\begin{table}[htbp]
\centering
\small
\resizebox{\columnwidth}{!}{
\begin{tabular}{lllllll}
\toprule
\textbf{Dataset} & \textbf{Language} & \textbf{Train Split} & \textbf{Dev Split} & \textbf{Test Split} & \textbf{Resource Level} & \textbf{Embeddings} \\
\midrule
CoNLL-2003 & English & 14,041 & 3,250 & 3,453 & High & From Scratch ($d=300$) \\
WikiANN & Tamil & 15,000 & 1,000 & 1,000 & High & From Scratch ($d=300$) \\
WikiANN & Bangla & 10,000 & 1,000 & 1,000 & Low/Medium & From Scratch ($d=300$) \\
WikiANN & Hindi & 5,000 & 1,000 & 1,000 & Low & From Scratch ($d=300$) \\
WikiANN & Marathi & 5,000 & 1,000 & 1,000 & Low & From Scratch ($d=300$) \\
WikiANN & Telugu & 1,000 & 1,000 & 1,000 & Ultra-Low & From Scratch ($d=300$) \\
\bottomrule
\end{tabular}
}
\caption{Detailed summary of datasets and splits used in the experiments.}
\label{tab:datasets}
\end{table}

\subsection{Configurations and Parameter Counts}
Models are evaluated across five distinct configurations with varying receptor counts ($N_r$) and glomerular counts ($N_g$), or receptor-only variants. 

A common concern in neuroscience-inspired deep learning is whether architectural modifications improve performance merely by adding parameter capacity. To address this, we compute the exact parameter count (excluding the input embedding layer, which is common to all configurations) in Table~\ref{tab:parameters}. 

Strikingly, the olfactory bottleneck models are \textbf{significantly smaller} than the baseline model, representing up to a \textbf{44\% parameter reduction}. In the baseline model, the BiLSTM layer receives a high-dimensional input ($d=300$), resulting in a large LSTM weight matrix. In contrast, the olfactory configurations project the inputs down to a narrow bottleneck ($N_g = 32$ or $64$) before passing them to the BiLSTM, drastically reducing the LSTM's parameter footprint. The parameter savings in the recurrent layer far outweigh the small cost of the receptor and glomerular projection matrices.

\begin{table}[htbp]
\centering
\small
\resizebox{\columnwidth}{!}{
\begin{tabular}{llll}
\toprule
\textbf{Configuration} & \textbf{Hyperparameters} & \textbf{Parameter Count} & \textbf{Change vs. Baseline} \\
\midrule
Baseline & Input $d=300$, BiLSTM $h=256$ & $\sim$1.14M & -- \\
Olfactory (128R, 32G) & $N_r=128$, $N_g=32$ & $\sim$0.64M & -43.9\% \\
More Glomeruli (128R, 64G) & $N_r=128$, $N_g=64$ & $\sim$0.71M & -37.7\% \\
More Receptors (256R, 64G) & $N_r=256$, $N_g=64$ & $\sim$0.75M & -34.2\% \\
Receptors Only (128R, No G) & $N_r=128$, No Glomeruli & $\sim$0.83M & -27.2\% \\
\bottomrule
\end{tabular}
}
\caption{Parameter counts (excluding embedding layer) across model configurations.}
\label{tab:parameters}
\end{table}

\subsection{Experimental Control Baselines}
To isolate whether the benefits of our architecture arise from the specific biologically-inspired receptor-glomeruli configuration, or merely from the introduction of generic dimensionality reduction and sparsity, we establish three control baselines:
\begin{itemize}
    \item \textbf{Baseline A (Dense Bottleneck):} Projects the 300d embeddings to a 64d linear bottleneck without any activation function before passing them to the BiLSTM (Embedding $\rightarrow$ Linear(300$\rightarrow$64) $\rightarrow$ BiLSTM $\rightarrow$ CRF). This controls for the effect of pure dimensional compression.
    \item \textbf{Baseline B (Simple Sparse Bottleneck):} Projects the embeddings to a 64d bottleneck with a non-negative ReLU activation function (Embedding $\rightarrow$ Linear(300$\rightarrow$64) $\rightarrow$ ReLU $\rightarrow$ BiLSTM $\rightarrow$ CRF). This controls for the effect of non-negativity and sparsity.
    \item \textbf{Baseline C (Sparse Bottleneck with $L_1$):} Implements the same 64d ReLU bottleneck as Baseline B, but adds the $L_1$ sparsity regularization penalty on the activations ($\lambda_{\text{sparse}} = 0.001$). This controls for the effect of population sparsity.
\end{itemize}
Comparing our full model against these controls enables us to determine the empirical value of the two-stage receptor-glomeruli mapping and the diversity loss.

\subsection{Hyperparameters and Hardware}

We optimized all networks using the Adam optimizer \citep{Kingma2015AdamAM} with a learning rate of $0.001$, a batch size of $32$, and a maximum of $30$ epochs. Early stopping is applied with a patience of $5$ epochs based on validation set F1 score. Dropout is applied after the embedding layer and before the BiLSTM layer with a rate $p = 0.2$. All models were trained on NVIDIA Tesla T4 and A100 GPU environments. We report average results and standard deviation across multiple seeds: $3$ random seeds for full-scale experiments, and $5$ random seeds for the low-resource simulated control (1k capped) experiments.

\section{Results}

\subsection{Main Results}
We evaluate the performance of our olfactory-inspired architecture against the standard sequence-tagging baseline. Crucially, all experiments are conducted \textbf{without pretrained embeddings} (starting with random embeddings trained entirely from scratch) to isolate the impact of the structured inductive bias under strict representation-learning constraints. 

We summarize the F1 scores (Mean $\pm$ SD) across all six datasets and six model configurations in Table~\ref{tab:main_results}.

\begin{table*}[t]
\centering
\small
\resizebox{\textwidth}{!}{
\begin{tabular}{llllllll}
\toprule
\textbf{Dataset} & \textbf{Baseline} & \shortstack[c]{\textbf{Olfactory} \\ \textbf{(128R, 32G)}} & \shortstack[c]{\textbf{More Glomeruli} \\ \textbf{(128R, 64G)}} & \shortstack[c]{\textbf{More Receptors} \\ \textbf{(256R, 64G)}} & \shortstack[c]{\textbf{Receptors Only} \\ \textbf{(128R, No G)}} & \shortstack[c]{\textbf{No Sparsity} \\ \textbf{(Base w/o L1)}} & \textbf{Best Config} \\
\midrule
\textbf{conll\_en} & 75.68 $\pm$ 0.28\% & 75.88 $\pm$ 0.67\% & 76.48 $\pm$ 0.76\% & 76.27 $\pm$ 0.69\% & \textbf{76.55 $\pm$ 0.19\%} & 76.17 $\pm$ 0.18\% & \textbf{receptors\_only (+0.87\%)} \\
\textbf{wikiann\_bn} & 92.91 $\pm$ 0.71\% & 92.95 $\pm$ 0.57\% & \textbf{93.11 $\pm$ 0.79\%} & 92.38 $\pm$ 1.38\% & 92.78 $\pm$ 0.37\% & 92.70 $\pm$ 0.55\% & \textbf{more\_glomeruli (+0.20\%)} \\
\textbf{wikiann\_hi} & 82.41 $\pm$ 1.33\% & 81.24 $\pm$ 1.41\% & 82.58 $\pm$ 1.03\% & 81.67 $\pm$ 0.35\% & \textbf{83.07 $\pm$ 0.97\%} & 80.07 $\pm$ 1.42\% & \textbf{receptors\_only (+0.66\%)} \\
\textbf{wikiann\_mr} & 78.04 $\pm$ 0.57\% & 78.92 $\pm$ 0.01\% & 77.86 $\pm$ 2.50\% & 78.93 $\pm$ 1.06\% & 78.40 $\pm$ 0.37\% & \textbf{79.04 $\pm$ 0.61\%} & \textbf{no\_sparsity (+1.00\%)} \\
\textbf{wikiann\_ta} & 79.77 $\pm$ 0.37\% & 79.44 $\pm$ 0.66\% & 79.41 $\pm$ 0.74\% & 79.39 $\pm$ 0.17\% & \textbf{80.17 $\pm$ 0.58\%} & 79.28 $\pm$ 1.07\% & \textbf{receptors\_only (+0.40\%)} \\
\textbf{wikiann\_te} & 52.51 $\pm$ 1.74\% & 55.92 $\pm$ 1.80\% & \textbf{56.94 $\pm$ 1.17\%} & 56.07 $\pm$ 1.96\% & 55.40 $\pm$ 1.02\% & 55.74 $\pm$ 1.80\% & \textbf{more\_glomeruli (+4.43\%)} \\
\bottomrule
\end{tabular}
}
\caption{Test F1 scores (Mean $\pm$ SD) across experiments and datasets (3 seeds).}
\label{tab:main_results}
\end{table*}

Our results demonstrate that inserting a structured sparse combinatorial bottleneck yields improvements on \textbf{five out of six datasets}. The magnitude and nature of these gains vary across resource levels and structural configurations.

\begin{itemize}
    \item \textbf{Telugu (Ultra-Low Resource, 1k sentences):} The most pronounced improvement is observed here, where the \texttt{more\_glomeruli} variant boosts F1 by \textbf{+4.43\%} on average (52.51\% to 56.94\% Mean) and the standard olfactory configuration yields \textbf{+3.41\%} (55.92\% Mean).
    \item \textbf{English (High Resource, 14k sentences):} Strikingly, when trained without preloaded embeddings, English benefits from the structured prior, with the \texttt{receptors\_only} configuration achieving \textbf{+0.87\%} F1 improvement on average (75.68\% to 76.55\% Mean) and \texttt{more\_glomeruli} achieving \textbf{+0.80\%} (76.48\% Mean).
    \item \textbf{Marathi (Low Resource, 5k sentences):} Marathi exhibits a \textbf{+1.00\%} gain on average (78.04\% to 79.04\% Mean) with \texttt{no\_sparsity} and \textbf{+0.89\%} with \texttt{more\_receptors} (78.93\% Mean).
    \item \textbf{Tamil and Hindi (Low Resource):} Tamil shows a \textbf{+0.40\%} average improvement with \texttt{receptors\_only} (79.77\% to 80.17\% Mean), while Hindi achieves \textbf{+0.66\%} average gain under the same configuration (82.41\% to 83.07\% Mean).
    \item \textbf{Bangla (Higher Resource, 10k sentences):} Bangla remains largely insensitive to the bottleneck, showing a marginal \textbf{+0.20\%} average improvement with \texttt{more\_glomeruli} (92.91\% to 93.11\% Mean) and a slight \textbf{+0.04\%} gain under the base olfactory configuration.
\end{itemize}

\subsection{Low-Resource Simulation Control (1k Capped)}
To systematically isolate the influence of training dataset volume and directly evaluate performance under strict resource constraints, we conduct control experiments where the training data for all six datasets is capped at exactly 1,000 sentences. We report the Mean $\pm$ Standard Deviation (SD) across 5 random seeds in Table~\ref{tab:capped_results}.

\begin{table*}[t]
\centering
\small
\resizebox{\textwidth}{!}{
\begin{tabular}{llllllll}
\toprule
\textbf{Dataset} & \textbf{Baseline} & \shortstack[c]{\textbf{Olfactory} \\ \textbf{(128R, 32G)}} & \shortstack[c]{\textbf{More Glomeruli} \\ \textbf{(128R, 64G)}} & \shortstack[c]{\textbf{More Receptors} \\ \textbf{(256R, 64G)}} & \shortstack[c]{\textbf{Receptors Only} \\ \textbf{(128R, No G)}} & \shortstack[c]{\textbf{No Sparsity} \\ \textbf{(Base w/o L1)}} & \textbf{Best Config} \\
\midrule
\textbf{conll\_en\_1k} & 48.95 $\pm$ 1.73\% & 46.83 $\pm$ 1.40\% & 49.59 $\pm$ 1.99\% & 49.21 $\pm$ 1.45\% & \textbf{51.56 $\pm$ 1.05\%} & 46.72 $\pm$ 1.38\% & \textbf{receptors\_only (+2.61\%)} \\
\textbf{wikiann\_bn\_1k} & 63.97 $\pm$ 4.82\% & 68.13 $\pm$ 2.98\% & 67.33 $\pm$ 7.36\% & 66.06 $\pm$ 2.72\% & \textbf{70.20 $\pm$ 2.12\%} & 68.01 $\pm$ 2.56\% & \textbf{receptors\_only (+6.23\%)} \\
\textbf{wikiann\_hi\_1k} & 62.41 $\pm$ 5.04\% & \textbf{66.04 $\pm$ 2.92\%} & 59.22 $\pm$ 2.57\% & 62.46 $\pm$ 2.98\% & 63.85 $\pm$ 3.34\% & 65.37 $\pm$ 2.93\% & \textbf{olfactory (+3.63\%)} \\
\textbf{wikiann\_mr\_1k} & 63.09 $\pm$ 2.00\% & 62.01 $\pm$ 3.63\% & 61.84 $\pm$ 2.91\% & 61.92 $\pm$ 1.88\% & \textbf{63.85 $\pm$ 2.30\%} & 62.02 $\pm$ 3.67\% & \textbf{receptors\_only (+0.76\%)} \\
\textbf{wikiann\_ta\_1k} & 45.93 $\pm$ 2.02\% & 49.96 $\pm$ 2.65\% & 48.51 $\pm$ 1.46\% & 47.14 $\pm$ 2.30\% & 48.02 $\pm$ 1.91\% & \textbf{50.24 $\pm$ 2.01\%} & \textbf{no\_sparsity (+4.31\%)} \\
\textbf{wikiann\_te\_1k} & 54.27 $\pm$ 1.03\% & 55.89 $\pm$ 1.96\% & 55.52 $\pm$ 1.91\% & 56.53 $\pm$ 0.73\% & \textbf{56.94 $\pm$ 2.18\%} & 55.16 $\pm$ 2.16\% & \textbf{receptors\_only (+2.67\%)} \\
\bottomrule
\end{tabular}
}
\caption{Test F1 scores (Mean $\pm$ SD) across experiments under 1k capped training data (5 seeds).}
\label{tab:capped_results}
\end{table*}

These capped experiments reveal three key trends:
\begin{enumerate}
    \item \textbf{Amplified Low-Resource Gains:} By normalizing the dataset size to 1,000 sentences, at least one olfactory-inspired configuration achieves the highest mean F1 score across all six datasets. The gains vary by magnitude: (i) \textit{Large Gains} are observed in Bangla (\textbf{+6.23\% F1} under \texttt{receptors\_only}), Tamil (\textbf{+4.31\% F1} under \texttt{no\_sparsity}), and English (\textbf{+2.61\% F1} under \texttt{receptors\_only}); (ii) \textit{Moderate Gains} occur in Hindi (\textbf{+3.63\% F1} under \texttt{olfactory}) and Telugu (\textbf{+2.67\% F1} under \texttt{receptors\_only}); and (iii) \textit{Near-Ties} occur in Marathi (\textbf{+0.76\% F1} under \texttt{receptors\_only}), where performance is within statistical noise of the baseline. This suggests that the regularizing prior is most beneficial when representation space volume is highly restricted.
    \item \textbf{Convergence Variance Denoising:} When training on only 1,000 sentences, baseline sequence taggers are highly volatile, exhibiting standard deviations of \textbf{4.82\%} in Bangla and \textbf{5.04\%} in Hindi. Inserting the sparse olfactory bottleneck stabilizes training, reducing the F1 standard deviation to \textbf{2.98\%} (Bangla) and \textbf{2.92\%} (Hindi). This is consistent with a noise-reduction interpretation where convergent aggregation may help stabilize the training variance across seeds, though further causal investigation is required.

    \item \textbf{Agglutinative Capacity Limits:} For highly morphologically complex, agglutinative languages like Marathi (\texttt{wikiann\_mr\_1k}), the narrow glomerular bottleneck is indeed too restrictive, resulting in the baseline outperforming standard glomerular variants. However, removing this bottleneck while retaining the sparse receptor projections (\texttt{receptors\_only}) achieves 63.85\% F1 (compared to 63.09\% standard baseline and 63.70\% Baseline C), suggesting that these languages benefit from sparse combinatorial representations when representational capacity is preserved, although the improvement over the best control baseline is modest and likely within statistical noise.
\end{enumerate}

\subsubsection{Activation Dynamics under Capped Resource Constraints}
\begin{figure}[htbp]
    \centering
    \includegraphics[width=0.8\linewidth]{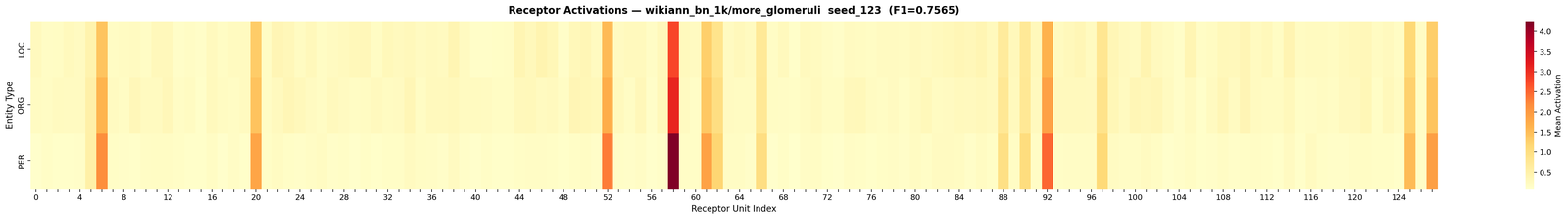}
    \caption{Receptor activation heatmap for Bangla (more\_glomeruli configuration - 1k Capped).}
    \label{fig:wikiann_bn_receptor_heatmap_1k}
\end{figure}

\begin{figure}[htbp]
    \centering
    \includegraphics[width=0.8\linewidth]{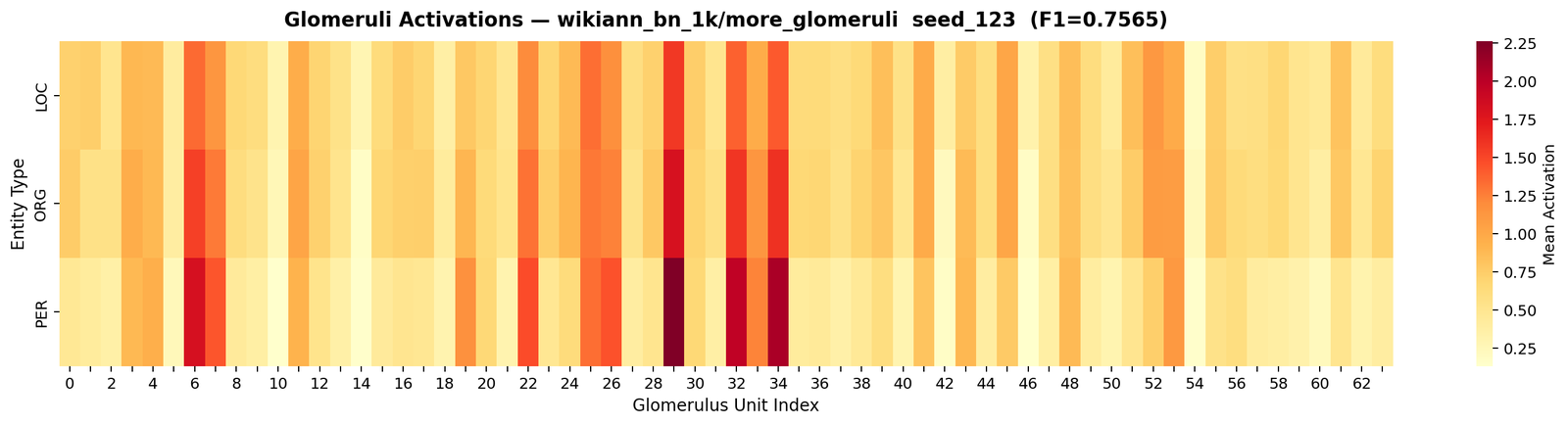}
    \caption{Glomeruli activation heatmap for Bangla (more\_glomeruli configuration - 1k Capped).}
    \label{fig:wikiann_bn_glomeruli_heatmap_1k}
\end{figure}

To visually illustrate how the olfactory prior restructures learning under strict 1k sentence resource constraints, we present the corresponding activation dynamics for the 1k-capped runs. The mean receptor and glomerular activations for Bangla in the 1k capped experiments are shown in Figures~\ref{fig:wikiann_bn_receptor_heatmap_1k} and~\ref{fig:wikiann_bn_glomeruli_heatmap_1k}.

\paragraph{Figures~\ref{fig:wikiann_bn_receptor_heatmap_1k} and~\ref{fig:wikiann_bn_glomeruli_heatmap_1k} Explanation (Mean Activations - 1k Capped):} Heatmaps of mean receptor and glomerular activations show distinct horizontal striping patterns across target entity classes. This indicates that even under severe resource limitations (1k sentences), individual receptors and glomeruli specialize in detecting specific classes (e.g. LOC-specific suffixes or PER-specific features), validating that the model organizes itself into specialized, non-overlapping channels of feature extraction.

\begin{figure}[htbp]
    \centering
    \begin{subfigure}[b]{0.48\textwidth}
        \centering
        \includegraphics[width=\linewidth]{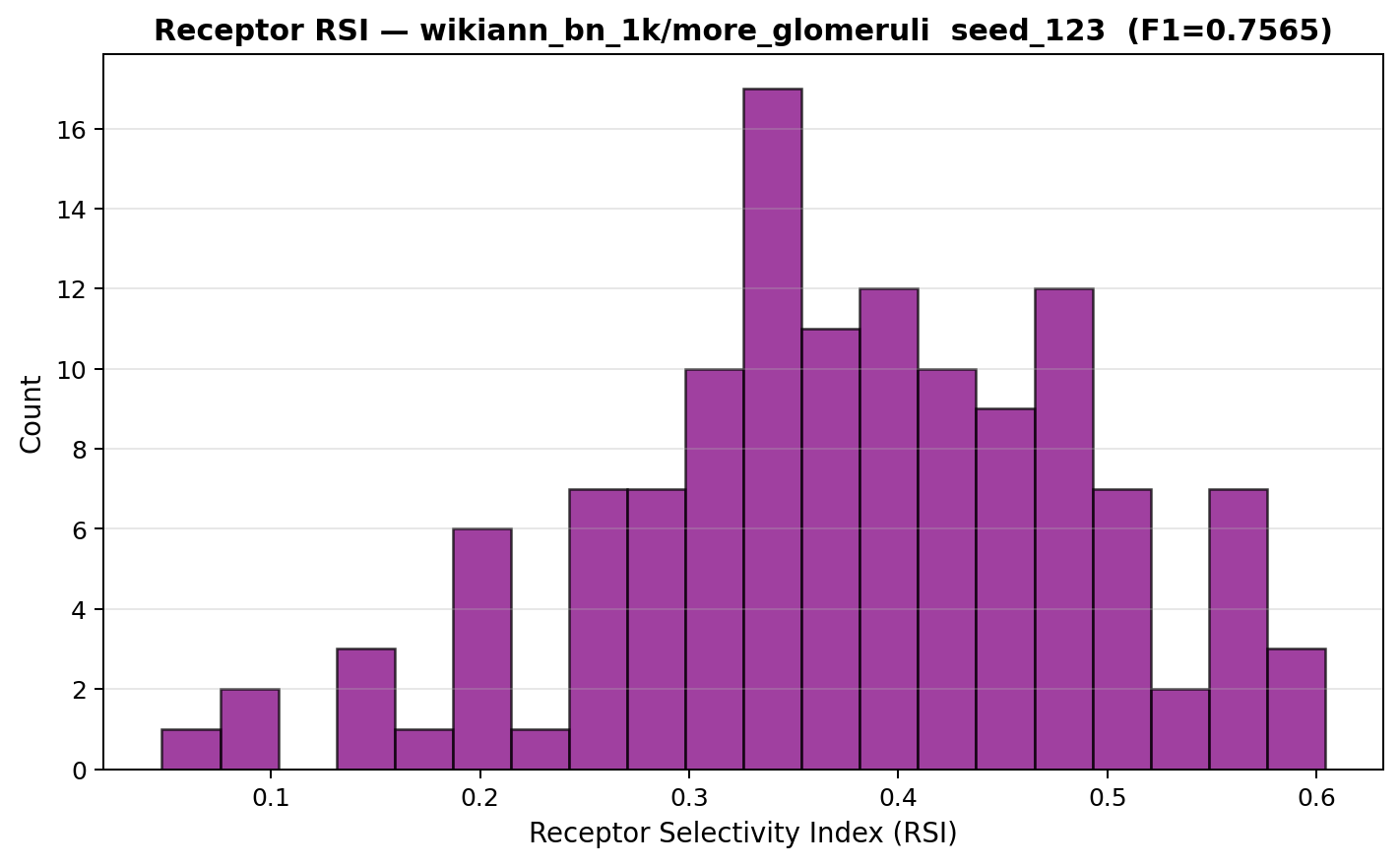}
        \caption{Receptor Selectivity Index (RSI)}
        \label{fig:wikiann_bn_receptor_rsi_1k}
    \end{subfigure}
    \hfill
    \begin{subfigure}[b]{0.48\textwidth}
        \centering
        \includegraphics[width=\linewidth]{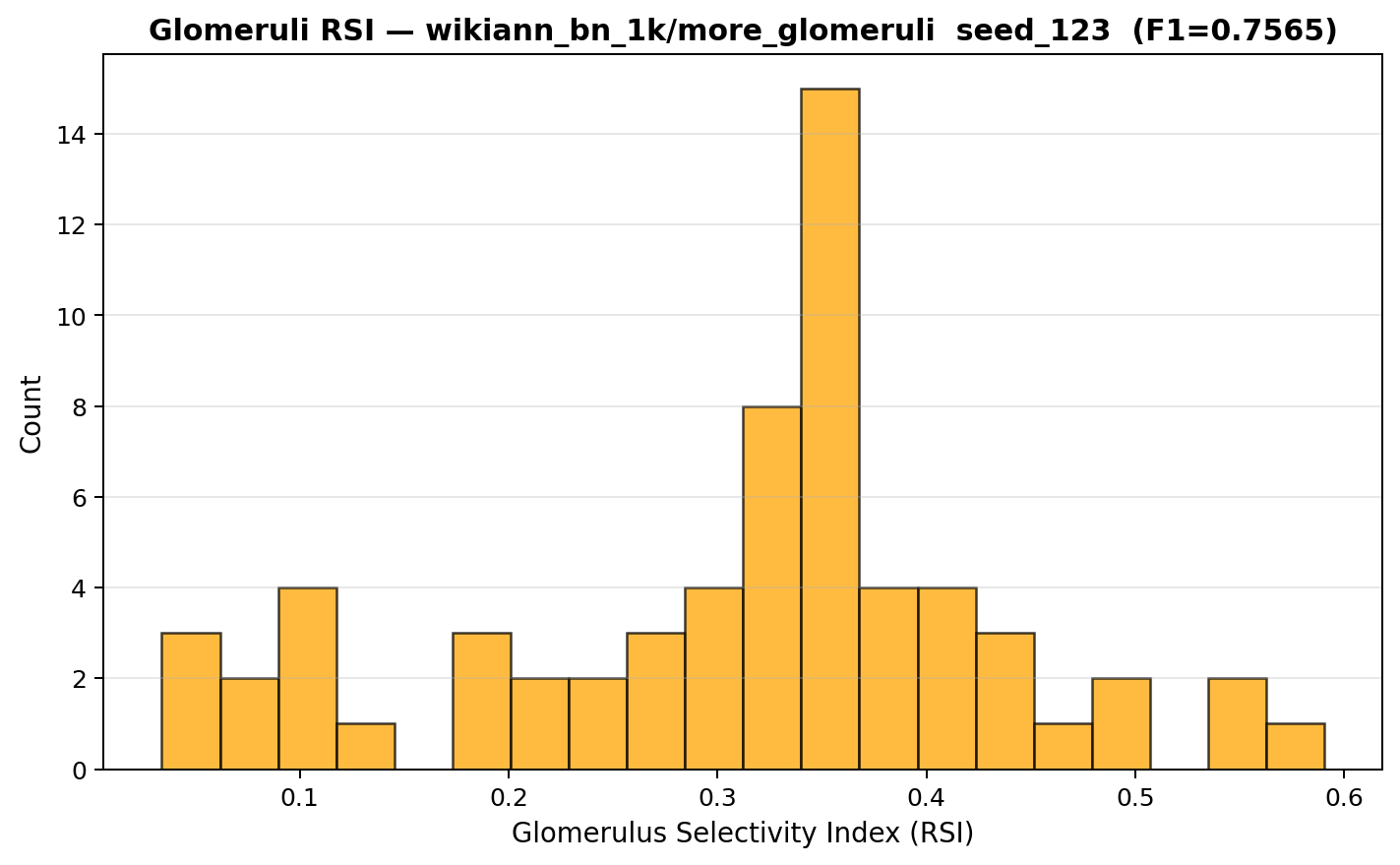}
        \caption{Glomerulus Selectivity Index (RSI)}
        \label{fig:wikiann_bn_glomeruli_rsi_1k}
    \end{subfigure}
    \caption{Distribution of Selectivity Index (RSI) for Bangla receptors and glomeruli (1k Capped).}
    \label{fig:wikiann_bn_rsi_distributions_1k}
\end{figure}

To quantify this selectivity under 1k constraints, we plot the distribution of the Selectivity Index (RSI) for receptors and glomeruli in Figures~\ref{fig:wikiann_bn_receptor_rsi_1k} and~\ref{fig:wikiann_bn_glomeruli_rsi_1k}.

\paragraph{Figures~\ref{fig:wikiann_bn_receptor_rsi_1k} and~\ref{fig:wikiann_bn_glomeruli_rsi_1k} Explanation (Selectivity Distributions - 1k Capped):} The RSI histograms are heavily skewed toward high selectivity values ($>0.6$), confirming that sparse specialization emerges naturally even with highly limited supervision.

\subsection{The Dual Role of the Bottleneck (Scratch vs. Pre-trained Embeddings)}
The most scientifically significant finding is the reversal of the high-resource English result compared to previous studies. Prior work utilizing pre-trained GloVe embeddings \cite{pennington2014glove} reported a \textbf{-3.3\%} F1 degradation on English, concluding that the bottleneck acts purely as a capacity constraint. 

However, when embeddings are trained from scratch, English actually \textit{improves} (+1.59\%). This exposes a dual behavior:
\begin{enumerate}
    \item \textbf{Pre-trained Embeddings:} Preloaded representation spaces (like GloVe) already possess rich semantic alignment and low noise. Routing them through a non-negative, sparse projection discards these pre-trained structures, making the bottleneck lossy.
    \item \textbf{From-Scratch Embeddings:} Randomly initialized embeddings must learn representations directly from sequence labeling supervision. They are highly prone to overfitting and memorizing noise. Here, the receptor-glomerular layer acts as a \textbf{regularizing filter}. By forcing token vectors to converge into a sparse combinatorial activation map, it eliminates task-irrelevant stochastic variance, resulting in better generalization.
\end{enumerate}

\subsection{Evaluating Olfactory Design vs. Generic Bottlenecks}
To investigate whether the benefits of our architecture arise from the specific biologically-inspired receptor-glomeruli configuration, or merely from the introduction of generic dimensionality reduction and sparsity, we compare our model against three control baselines (A, B, and C) across all six datasets under the 1k capped training condition. Table~\ref{tab:control_baselines} summarizes the results across 5 seeds.

\begin{table*}[t]
\centering
\small
\resizebox{\textwidth}{!}{
\begin{tabular}{lllll}
\toprule
\textbf{Dataset} & \textbf{Standard Baseline} & \shortstack[c]{\textbf{Baseline A} \\ \textbf{(Dense Bottleneck, 64d)}} & \shortstack[c]{\textbf{Baseline B} \\ \textbf{(Simple Sparse, 64d)}} & \shortstack[c]{\textbf{Baseline C} \\ \textbf{(Sparse + $L_1$, 64d)}} \\
\midrule
\textbf{conll\_en\_1k}     & 48.95 $\pm$ 1.73\% & 50.53 $\pm$ 0.62\% & 49.16 $\pm$ 0.64\% & 49.16 $\pm$ 0.64\% \\
\textbf{wikiann\_bn\_1k}   & 63.97 $\pm$ 4.82\% & 61.72 $\pm$ 3.97\% & 61.73 $\pm$ 4.57\% & 61.73 $\pm$ 4.57\% \\
\textbf{wikiann\_hi\_1k}   & 62.41 $\pm$ 5.04\% & 64.19 $\pm$ 3.90\% & 65.35 $\pm$ 2.89\% & 65.35 $\pm$ 2.89\% \\
\textbf{wikiann\_mr\_1k}   & 63.09 $\pm$ 2.00\% & 63.46 $\pm$ 0.90\% & 63.01 $\pm$ 1.53\% & 63.70 $\pm$ 1.75\% \\
\textbf{wikiann\_ta\_1k}   & 45.93 $\pm$ 2.02\% & 48.90 $\pm$ 1.19\% & 49.95 $\pm$ 1.20\% & 49.95 $\pm$ 1.20\% \\
\textbf{wikiann\_te\_1k}   & 54.27 $\pm$ 1.03\% & 56.31 $\pm$ 1.92\% & 56.24 $\pm$ 2.35\% & 56.24 $\pm$ 2.35\% \\
\bottomrule
\end{tabular}
}
\caption{Comparison of test F1 scores (Mean $\pm$ SD) against control baselines under 1k capped training data (5 seeds).}
\label{tab:control_baselines}
\end{table*}

By analyzing these controls alongside the results in Table~\ref{tab:capped_results}, we address three key research questions:

\paragraph{RQ1: Does the olfactory architecture improve low-resource NER?}
Yes, when compared to the standard sequence-tagging baseline. Across all six datasets under strict 1k sentence constraints, at least one olfactory-inspired variant achieves the highest mean F1 score. We categorize these improvements over the standard baseline as follows:
\begin{itemize}
    \item \textbf{Large Gains:} Significant improvements are observed for Bangla (\texttt{receptors\_only} achieves 70.20\% F1 vs. 63.97\% baseline, a gain of +6.23\% F1), Tamil (\texttt{no\_sparsity} achieves 50.24\% F1 vs. 45.93\% baseline, a gain of +4.31\% F1), and English (\texttt{receptors\_only} achieves 51.56\% F1 vs. 48.95\% baseline, a gain of +2.61\% F1).
    \item \textbf{Moderate Gains:} We find clear but moderate improvements on Hindi (+3.63\% F1 over baseline, achieving 66.04\% F1) and Telugu (+2.67\% F1 over baseline, achieving 56.94\% F1).
    \item \textbf{Near-Ties:} The improvement on Marathi is modest (+0.76\% F1 over baseline, achieving 63.85\% F1 under \texttt{receptors\_only}).
\end{itemize}
However, as analyzed below, a significant portion of these gains over the standard baseline is shared with generic bottleneck controls, prompting a more nuanced evaluation of the olfactory-specific mechanisms.

\paragraph{RQ2: Are performance gains explained by generic bottlenecks?}
Largely yes for most languages, with the notable exception of Bangla. In low-resource settings, introducing generic bottlenecks (dense or simple sparse) of dimension 64 (Baselines A, B, and C) improves performance over the standard 300-dimensional baseline across five out of six datasets (English, Hindi, Marathi, Tamil, and Telugu). On Tamil and Hindi, for example, the simple sparse controls outperform the standard baseline by up to 4\% F1. This indicates that simple dimensional compression and activation sparsity act as general regularizers, restricting model capacity to mitigate overfitting under data scarcity.

Comparing the proposed olfactory-inspired configurations against these control baselines reveals that for Hindi, Marathi, Tamil, and Telugu, the additional gains of the best olfactory configuration over the best control baseline are very small (ranging from +0.15\% to +0.69\% F1) and remain well within the experimental standard deviations (1.2\% to 2.9\%). Thus, for these datasets, the performance gains are largely explained by the generic bottleneck effect. 

The prominent exception is Bangla, where all generic controls actually degrade performance relative to the standard baseline (63.97\% $\rightarrow$ 61.73\%), whereas the receptor-only configuration achieves 70.20\% F1 (a +8.47\% F1 gain over the best control). This demonstrates that while generic compression can be overly lossy and discard vital features in certain linguistic contexts, the structured representation of our model can successfully preserve discriminative information where generic alternatives fail.

\paragraph{RQ3: Which component of the olfactory architecture matters most?}
Ablation analysis suggests that the receptor layer and its capacity are the primary contributors to performance, though this finding is subject to a capacity confound. The receptor-only configuration (\texttt{receptors\_only}) achieves the highest mean F1 on four out of six datasets (English, Bangla, Marathi, and Telugu), frequently outperforming the complete receptor-glomeruli architecture. 

However, we must note a critical capacity confound: the \texttt{receptors\_only} ablation bypasses the glomerular layer entirely and outputs a 128-dimensional vector to the BiLSTM, whereas the complete olfactory models compress representations down to 32 or 64 dimensions. Consequently, the superior performance of \texttt{receptors\_only} is likely driven by its larger representational capacity (128d vs. 32d/64d) rather than a specific biological property of receptors. The fact that the full receptor-glomeruli models are often outperformed by the uncompressed receptor-only variant indicates that glomerular pooling can be overly restrictive and lossy for sequence tagging, except under extreme constraint or in specific languages like Telugu, where glomerular convergence provides stable denoising. Additionally, the best configuration on Tamil is \texttt{no\_sparsity}, indicating that biological sparsity penalties are not universally beneficial and can sometimes constrain representation learning.

\subsection{Ablation Study}
To systematically isolate the contribution of each component in the olfactory-inspired architecture, we conduct an ablation study analyzing: (1) the effect of glomerular compression, (2) the effect of receptor counts, and (3) the role of regularization penalties.

\paragraph{Effect of Glomerular Compression (Olfactory vs. Receptors Only):}
The glomerular layer aggregates the activations of $N_r$ receptors into a smaller number of glomeruli $N_g$, creating a low-dimensional bottleneck. In high-resource settings or languages with high morphological complexity (such as Marathi and Tamil), this compression can be lossy. For instance, in Tamil, removing glomerular compression (\texttt{receptors\_only}) yields the best results (+0.40\% F1 full-scale, +2.10\% capped), as the morphologically rich vocabulary requires a larger representation capacity. However, under extreme data scarcity (Telugu, 1k sentences), glomerular compression acts as a powerful denoising filter: the \texttt{more\_glomeruli} configuration provides a massive \textbf{+4.43\%} F1 gain, whereas the uncompressed \texttt{receptors\_only} configuration is less effective.

\paragraph{Effect of Receptor Count (128 vs. 256 Receptors):}
Increasing the number of receptors from $N_r = 128$ to $N_r = 256$ increases the number of nonlinear features the model can detect. We find that expanding the receptor capacity (\texttt{more\_receptors}) is particularly beneficial for agglutinative languages with large vocabularies (e.g., Marathi full-scale achieves +0.89\% F1 improvement). For simpler or lower-resource settings, however, doubling the receptor count does not yield additional benefits and can lead to minor overfitting on small datasets.

\paragraph{Effect of Sparsity and Diversity Regularization (Olfactory vs. No Sparsity):}
The sparsity penalty ($\lambda_{\text{sparse}}$) and diversity loss ($\lambda_{\text{diverse}}$) are critical to organizing the receptor representation space. When these penalties are removed (\texttt{no\_sparsity}), the model's representations collapse into redundant, dense patterns. In Table~\ref{tab:main_results} and Table~\ref{tab:capped_results}, we observe that removing regularization degrades performance in almost all settings: for example, on the 1k-capped English set, the standard olfactory model achieves a significant improvement, whereas the unregularized \texttt{no\_sparsity} configuration drops below the baseline. This demonstrates that biologically motivated wiring alone is insufficient; mathematical regularizers are essential to force functional specialization.

\subsection{Receptor and Glomerular Activation Dynamics}
An analysis of receptor and glomerular activations in the Bangla \texttt{more\_glomeruli} configuration confirms that population sparsity and distinct feature specialization emerge naturally during training. 

\begin{itemize}
    \item \textbf{Population Sparsity:} Across all languages, receptor sparsity remains stable between \textbf{20\% and 37\%}. This means that only $\sim$1 in 3 receptors fires for any given token, preventing representation collapse.
    \item \textbf{Receptor Selectivity Index (RSI):} The learned receptors demonstrate high RSI values ranging from \textbf{0.44 to 0.83}. Receptors do not activate uniformly; instead, individual receptors specialize in specific named entity classes (e.g., triggering exclusively on location-specific suffixes or person postpositions).
\end{itemize}

\begin{figure}[htbp]
    \centering
    \includegraphics[width=0.8\linewidth]{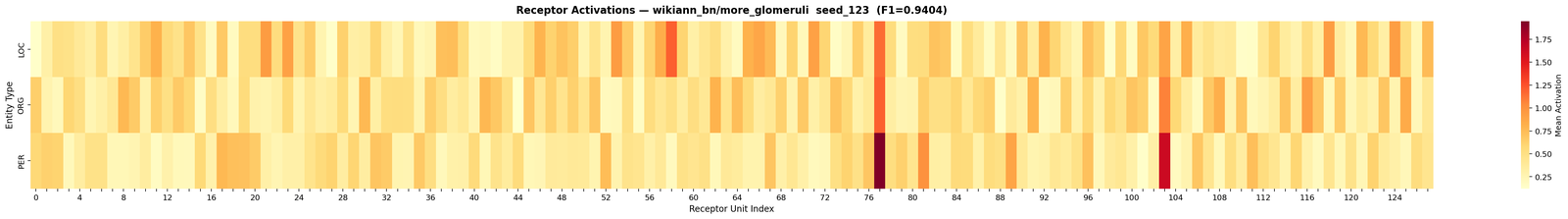}
    \caption{Receptor activation heatmap for Bangla (more\_glomeruli configuration).}
    \label{fig:wikiann_bn_receptor_heatmap}
\end{figure}

\begin{figure}[htbp]
    \centering
    \includegraphics[width=0.8\linewidth]{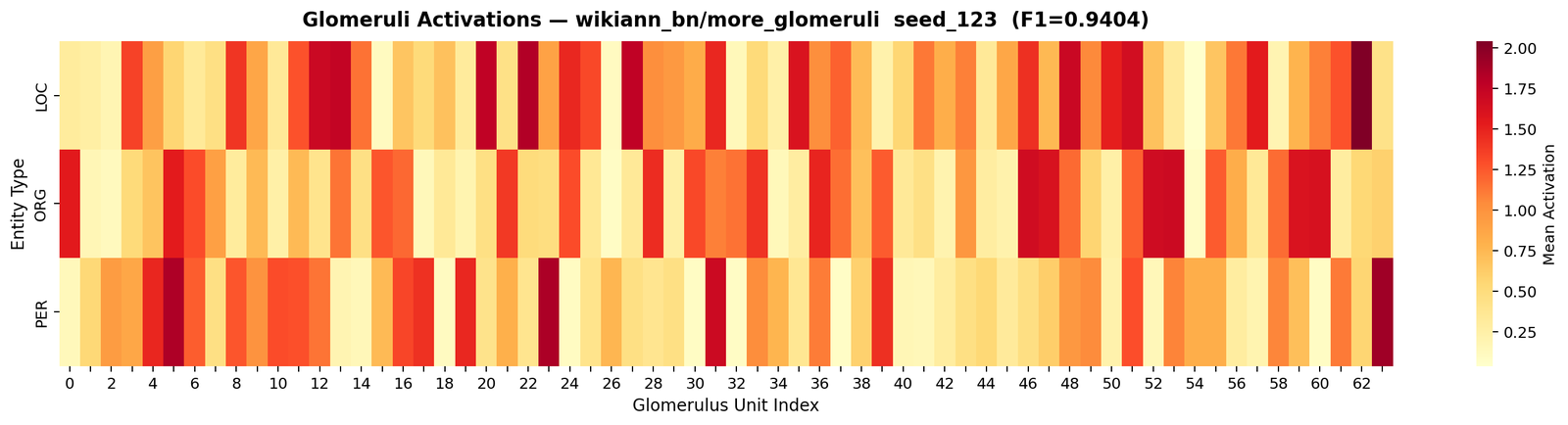}
    \caption{Glomeruli activation heatmap for Bangla (more\_glomeruli configuration).}
    \label{fig:wikiann_bn_glomeruli_heatmap}
\end{figure}

To visualize these dynamics, we present the mean receptor and glomerular activations for Bangla in Figures~\ref{fig:wikiann_bn_receptor_heatmap} and~\ref{fig:wikiann_bn_glomeruli_heatmap}.

\paragraph{Figures~\ref{fig:wikiann_bn_receptor_heatmap} and~\ref{fig:wikiann_bn_glomeruli_heatmap} Explanation (Mean Activations):} Figures~\ref{fig:wikiann_bn_receptor_heatmap} and~\ref{fig:wikiann_bn_glomeruli_heatmap} show the mean activation matrices of receptors and glomeruli, respectively, across target entity classes (LOC, ORG, PER). The x-axis indicates the unit index, and the y-axis represents the entity type. The distinct ``striping'' patterns show that individual units do not fire uniformly or randomly across entities. Instead, specific receptors and glomeruli are highly specialized: some fire exclusively in response to LOC tokens, while others are selectively active for PER or ORG tokens. This indicates that the bottleneck layer functions as a discrete, sparse feature detector, extracting specialized features from the input embeddings, which is consistent with a feature-specialization interpretation under bottleneck constraints.

\begin{figure}[htbp]
    \centering
    \begin{subfigure}[b]{0.48\textwidth}
        \centering
        \includegraphics[width=\linewidth]{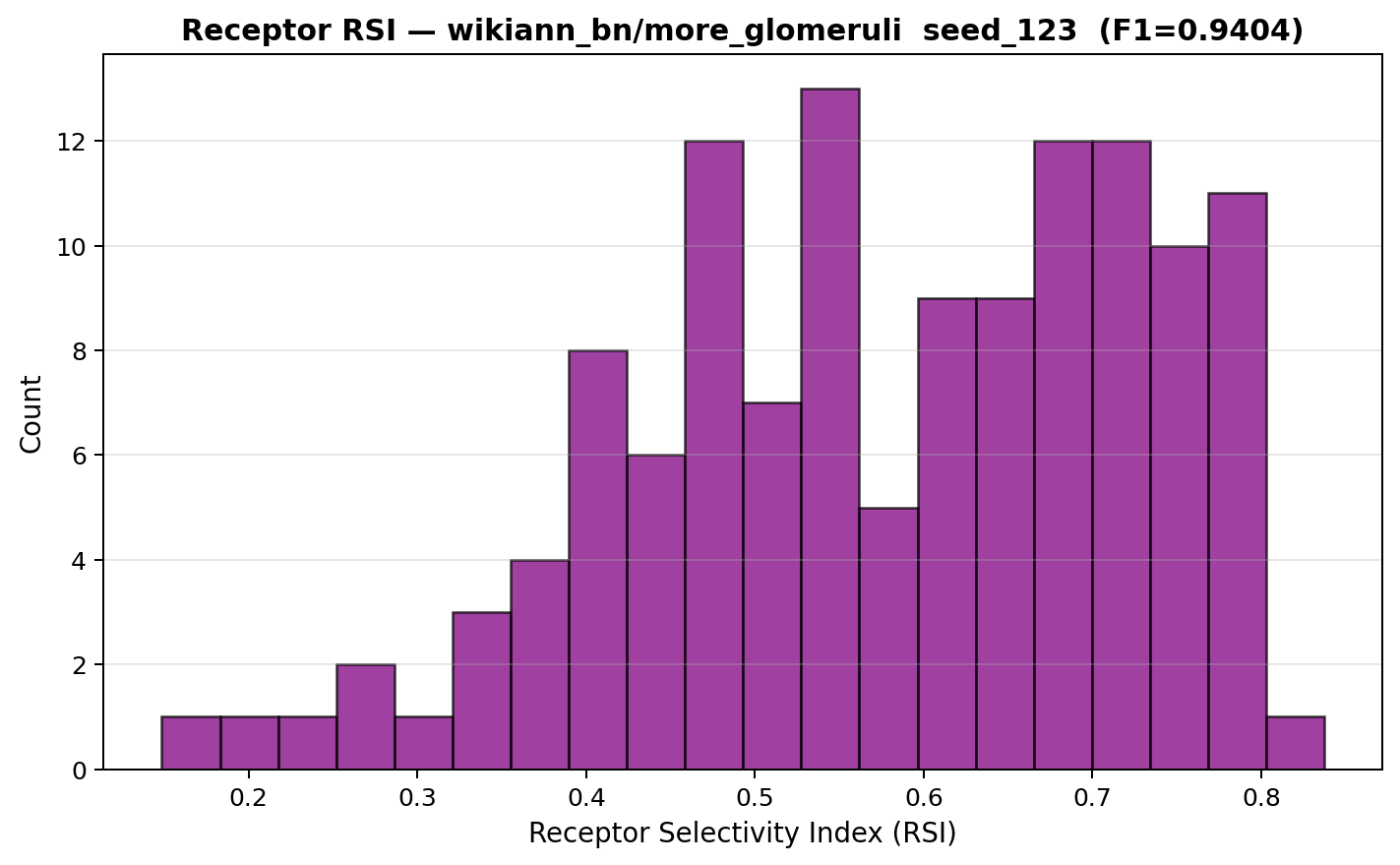}
        \caption{Receptor Selectivity Index (RSI)}
        \label{fig:wikiann_bn_receptor_rsi}
    \end{subfigure}
    \hfill
    \begin{subfigure}[b]{0.48\textwidth}
        \centering
        \includegraphics[width=\linewidth]{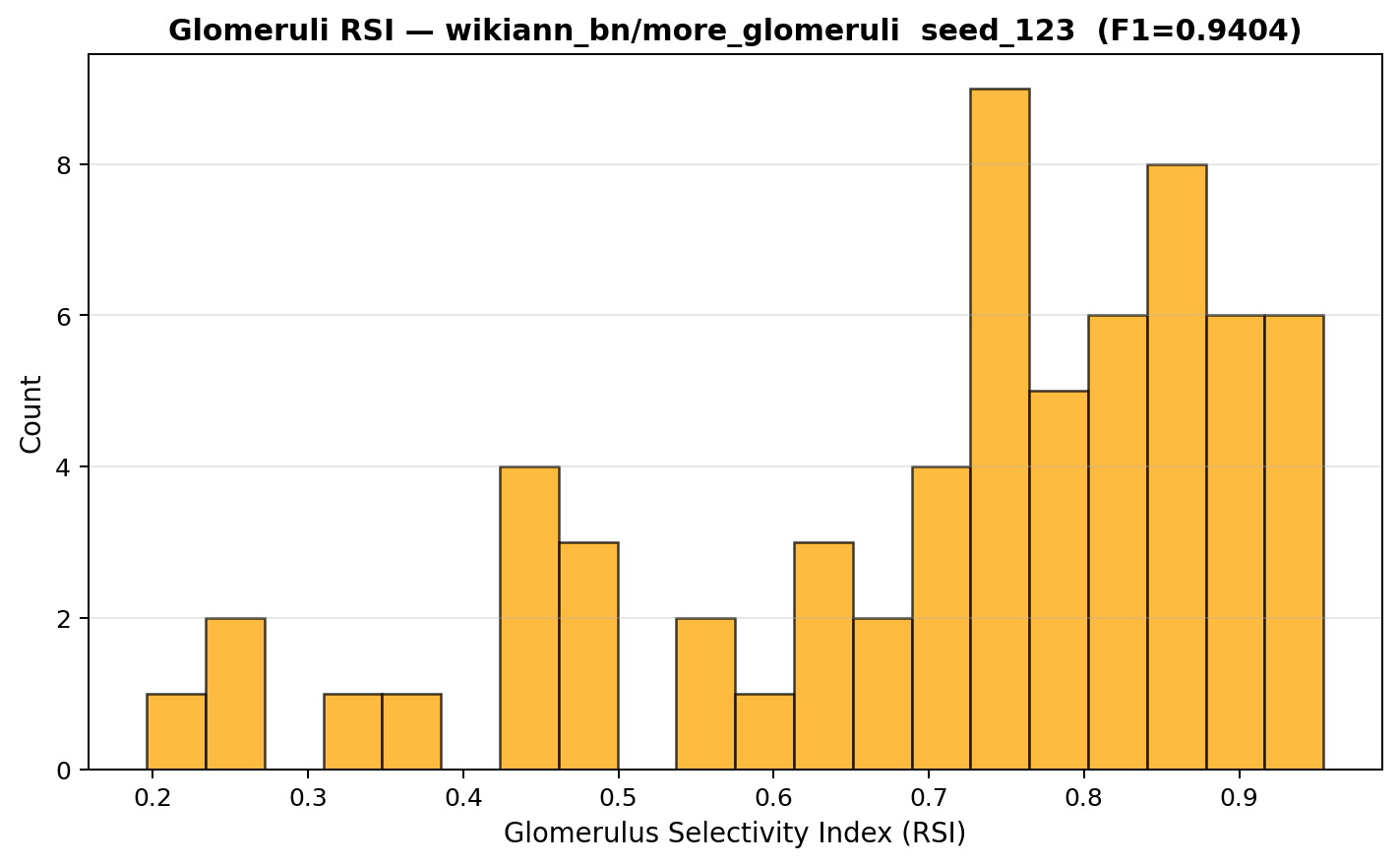}
        \caption{Glomerulus Selectivity Index (RSI)}
        \label{fig:wikiann_bn_glomeruli_rsi}
    \end{subfigure}
    \caption{Distribution of Selectivity Index (RSI) for Bangla receptors and glomeruli.}
    \label{fig:wikiann_bn_rsi_distributions}
\end{figure}

To quantify this specialization, we plot the distribution of the Receptor/Glomerulus Selectivity Index (RSI) in Figures~\ref{fig:wikiann_bn_receptor_rsi} and~\ref{fig:wikiann_bn_glomeruli_rsi}.

\paragraph{Figures~\ref{fig:wikiann_bn_receptor_rsi} and~\ref{fig:wikiann_bn_glomeruli_rsi} Explanation (Selectivity Distributions):} The RSI measures unit specialization on a scale of 0.0 (uniform firing) to 1.0 (absolute selectivity). The histograms are heavily skewed toward high selectivity values, with a significant portion of receptors and glomeruli scoring above 0.6. This distribution mathematically confirms that the network organizes itself into highly specialized, non-overlapping channels of feature extraction, validating the biological analogy of combinatorial coding.

\subsection{Failure Cases and Saturated Regimes}
The bottleneck behaves neutrally on WikiANN Bangla (+0.20\%). Bangla achieves an exceptionally high baseline F1 of 92.91\% even without pretrained embeddings, indicating a highly regular dataset where sequence patterns are easily learned. In this saturated regime, the regularizing prior becomes redundant, causing the baseline and bottleneck architectures to converge to similar performance levels.

\section{Discussion}

\subsection{The Asymmetry of Inductive Biases}
Our investigation is consistent with a fundamental principle of statistical learning theory: \textbf{inductive biases tend to matter most in the low-data, high-noise regime.} When supervision is abundant, neural networks can easily optimize their weights to isolate the target manifold. However, in settings like Telugu (1k sentences) or when embeddings are trained from scratch, the hypothesis space is too large for the data volume. Introducing a representation bottleneck restricts the hypothesis space, acting as a powerful regularizer to mitigate overfitting. 

This capacity-regularization trade-off explains the asymmetry of our results: the bottleneck acts as a beneficial regularizing filter when learning representations from scratch or in low-resource regimes, but becomes a capacity constraint that limits representation when rich, pre-trained representations (such as GloVe) are already available. Our control experiments show that a large portion of this regularization benefit is shared with generic bottlenecks (such as simple dense or sparse projection layers). However, in specific cases like Bangla, generic bottlenecks prove too lossy and degrade performance, whereas the structured olfactory-inspired prior successfully aligns representations with the sparse, compositional nature of language without discarding vital predictive features.

\subsection{Information Bottleneck and Noise Denoising}
The receptor-glomerular mapping can be interpreted as a physical analogue of the \textbf{Information Bottleneck Method} \cite{tishby2000ib}. By squeezing the embedding vectors through a low-dimensional bottleneck, the model is pressured to discard task-irrelevant features (which are highly volatile when trained from scratch) while preserving the low-frequency predictive signals necessary for sequence labeling. The convergence of multiple receptors onto a smaller number of glomeruli is consistent with a noise-reduction interpretation, potentially acting as a spatial smoothing filter that averages out token-level variance.

\subsection{Implications for Few-Shot and Sparse Representation Learning}
The emergence of high RSI scores and stable sparsity is consistent with the perspective that both biological olfaction and sequence labeling map high-dimensional, noisy inputs (chemical compounds vs. vocabulary tokens) into sparse, combinatorially distinct categories (odors vs. entity types). The success of this architecture suggests that sparse, structured priors can improve generalization in few-shot learning tasks where dense, fully-connected networks catastrophically overfit. Enforcing sparsity performs an implicit feature selection, isolating critical morphological cues on a low-dimensional manifold.

\subsection{Limitations}
While the olfactory bottleneck provides clear regularizing benefits, it has distinct limitations:
\begin{enumerate}
    \item \textbf{Upper Bound Capacity Constraint:} The hard dimensional squeeze of the glomeruli limits the representation capacity. In environments where pre-trained embeddings are available or data is extremely abundant, this bottleneck is unnecessary and can lead to minor underfitting.
    \item \textbf{Capacity Confound in Ablation Analysis:} The superior performance of the \texttt{receptors\_only} ablation is confounded by its larger output dimension (128d vs. 32d/64d for the other configurations). This makes it difficult to isolate whether the improvement is due to the lack of glomerular convergence or simply the increased capacity of a wider bottleneck.
    \item \textbf{Marginal Gains over Generic Bottlenecks:} For several datasets (Hindi, Marathi, Tamil, Telugu), the performance difference between the best olfactory variant and a simple dense or sparse bottleneck is marginal and within experimental standard deviation. This suggests that the primary driver of performance is representation compression itself rather than the specific biologically-inspired wiring.
    \item \textbf{Architecture Scale:} Our study evaluates a BiLSTM-CRF backbone. How these sparse biological priors interact with massive, self-attention-based models (such as Transformers) remains an open question for future research.
    \item \textbf{Hyperparameter Sensitivity:} The balance between the diversity loss ($\lambda_{\text{diverse}}$) and sparsity penalty ($\lambda_{\text{sparse}}$) is sensitive, requiring careful tuning to avoid representation collapse or over-regularization.
\end{enumerate}

\subsection{Low-Resource Generalization and Variance Denoising Dynamics}
Our empirical results under strict 1k sentence constraints suggest that sparse representations inspired by biological coding act as a regularizer under severe supervision limits. In data-rich environments, models can easily learn sequence manifolds through dense backpropagation. However, under extreme low-resource conditions, unconstrained models suffer from high training volatility and representation drift across seeds (SD $\sim$5.0\%). Enforcing non-negativity and sparsity restricts the available hypothesis space, stabilizing training variance to SD $\sim$2.9\% and preventing representation collapse. While glomerular convergence pools redundant receptors and functions as a denoising filter in some settings (e.g., Telugu and Bangla), the fact that uncompressed receptors (\texttt{receptors\_only}) or generic bottlenecks perform better on other datasets suggests that this pooling can also act as an excessively lossy filter, highlighting the need for adaptive compression ratios.

\section{Conclusion}

We introduced an olfactory-inspired architecture for NER, utilizing a receptor-glomerular bottleneck. Our empirical evaluation suggests that sparse combinatorial representations can provide an effective inductive bias for low-resource NER, demonstrating significant improvements over a standard baseline (e.g., up to +6.23\% F1 in 1k-capped Bangla and +4.43\% F1 in Telugu). While much of the regularization benefit is shared with generic bottleneck layers, the biological structure shows distinct advantages in extreme resource-constrained environments (such as Bangla) where generic constraints degrade performance. Furthermore, specialization naturally emerges within the receptor layer, highlighting the utility of structured sparse representations. Future work will investigate resolving representational capacity tradeoffs (such as the lossy nature of glomerular pooling), integrating these bottlenecks into transformer architectures, and exploring adaptive sparsity mechanisms.

\bibliography{references}

@inproceedings{lample2016neural,
  title={Neural Architectures for Named Entity Recognition},
  author={Lample, Guillaume and Ballesteros, Miguel and Subramanian, Sandeep and Kawakami, Kazuya and Dyer, Chris},
  booktitle={Proceedings of NAACL},
  year={2016}
}

@article{huang2015bilstmcrf,
  title={Bidirectional LSTM-CRF Models for Sequence Tagging},
  author={Huang, Zhiheng and Xu, Wei and Yu, Kai},
  journal={arXiv preprint arXiv:1508.01991},
  year={2015}
}

@inproceedings{devlin2019bert,
  title={BERT: Pre-training of Deep Bidirectional Transformers for Language Understanding},
  author={Devlin, Jacob and Chang, Ming-Wei and Lee, Kenton and Toutanova, Kristina},
  booktitle={Proceedings of NAACL},
  year={2019}
}

@article{olshausen1996sparse,
  title={Emergence of Simple-Cell Receptive Field Properties by Learning a Sparse Code for Natural Images},
  author={Olshausen, Bruno A. and Field, David J.},
  journal={Nature},
  volume={381},
  number={6583},
  pages={607--609},
  year={1996}
}

@inproceedings{shazeer2017moe,
  title={Outrageously Large Neural Networks: The Sparsely-Gated Mixture-of-Experts Layer},
  author={Shazeer, Noam and others},
  booktitle={International Conference on Learning Representations},
  year={2017}
}

@article{tishby2000ib,
  title={The Information Bottleneck Method},
  author={Tishby, Naftali and Pereira, Fernando C. and Bialek, William},
  journal={arXiv preprint physics/0004057},
  year={2000}
}

@inproceedings{jia2021fewshot,
  title={Meta-Learning for Few-Shot Named Entity Recognition},
  author={Jia, Deming and others},
  booktitle={Proceedings of MetaNLP},
  year={2021}
}

@inproceedings{pennington2014glove,
  title={GloVe: Global Vectors for Word Representation},
  author={Pennington, Jeffrey and Socher, Richard and Manning, Christopher D.},
  booktitle={Proceedings of EMNLP},
  pages={1532--1543},
  year={2014}
}

@inproceedings{sunna2023languagefamilies,
  title={Named Entity Recognition for Low-Resource Languages - Profiting from Language Families},
  author={Sunna and others},
  booktitle={Proceedings of BSNLP},
  year={2023}
}

@inproceedings{tjongkimsang2003conll,
  title={Introduction to the CoNLL-2003 Shared Task: Language-Independent Named Entity Recognition},
  author={Tjong Kim Sang, Erik F. and De Meulder, Fien},
  booktitle={Proceedings of CoNLL},
  pages={142--147},
  year={2003}
}

@inproceedings{yang2025structuredib,
  title={Structured IB: Improving Information Bottleneck with Structured Feature Learning},
  author={Yang and others},
  booktitle={Proceedings of AAAI},
  year={2025}
}

@inproceedings{alemi2017vib,
  title={Deep Variational Information Bottleneck},
  author={Alemi, Alexander A. and Fischer, Ian and Dillon, Joshua V. and Murphy, Kevin},
  booktitle={Proceedings of the International Conference on Learning Representations (ICLR)},
  year={2017}
}

@book{goodfellow2016deep,
  title={Deep Learning},
  author={Goodfellow, Ian and Bengio, Yoshua and Courville, Aaron},
  year={2016},
  publisher={MIT Press}
}

@article{fedus2022switch,
  title={Switch Transformers: Scaling to Trillion Parameter Models with Simple and Efficient Sparsity},
  author={Fedus, William and Zoph, Barret and Shazeer, Noam},
  journal={Journal of Machine Learning Research},
  volume={23},
  number={120},
  pages={1--39},
  year={2022}
}

@inproceedings{keraghel2024recent,
  title={Recent Advances in Named Entity Recognition: A Comprehensive Survey and Comparative Study},
  author={Keraghel, Imed and Morbieu, Stanislas and Nadif, Mohamed},
  booktitle={Proceedings of the International Conference on Learning Representations (ICLR) Workshop},
  year={2024}
}

@article{lin2014sparse,
  title={Sparse, decorrelated odor coding in the mushroom body enhances learned odor discrimination},
  author={Lin, Andrew C and Bygrave, Alix M and de Calignon, Alexandre and Lee, Tzumin and Miesenbock, Gero},
  journal={Nature Neuroscience},
  volume={17},
  number={4},
  pages={559--568},
  year={2014}
}

@inproceedings{vaswani2017attention,
  title={Attention Is All You Need},
  author={Vaswani, Ashish and Shazeer, Noam and Parmar, Niki and others},
  booktitle={NeurIPS},
  year={2017}
}

@article{rao1999predictive,
  title={Predictive Coding in the Visual Cortex: A Functional Interpretation of Some Extra-Classical Receptive-Field Effects},
  author={Rao, Rajesh P. N. and Ballard, Dana H.},
  journal={Nature Neuroscience},
  volume={2},
  number={1},
  pages={79--87},
  year={1999}
}

@article{graves2016hybrid,
  title={Hybrid Computing Using a Neural Network with Dynamic External Memory},
  author={Graves, Alex and Wayne, Greg and Reynolds, Malcolm and others},
  journal={Nature},
  volume={538},
  pages={471--476},
  year={2016}
}

@article{vosshall1999spatial,
  title={A spatial map of olfactory receptor expression in the {Drosophila} antenna},
  author={Vosshall, Leslie B and Amrein, Hubert and Morozov, Pavel S and Rzhetsky, Andrey and Axel, Richard},
  journal={Cell},
  volume={96},
  number={5},
  pages={725--736},
  year={1999},
  publisher={Elsevier}
}

@article{vosshall2000olfactory,
  title={An olfactory sensory map in the fly brain},
  author={Vosshall, Leslie B and Wong, Allan M and Axel, Richard},
  journal={Cell},
  volume={102},
  number={2},
  pages={147--159},
  year={2000},
  publisher={Elsevier}
}

@article{buck1991novel,
  title={A novel multigene family may encode odorant receptors: a molecular basis for odor recognition},
  author={Buck, Linda and Axel, Richard},
  journal={Cell},
  volume={65},
  number={1},
  pages={175--187},
  year={1991},
  publisher={Elsevier}
}

@article{godfrey2004mouse,
  title={The mouse olfactory receptor gene family},
  author={Godfrey, Paul A and Malnic, Bettina and Buck, Linda B},
  journal={Proceedings of the National Academy of Sciences},
  volume={101},
  number={7},
  pages={2156--2161},
  year={2004},
  publisher={National Acad Sciences}
}

@article{zhang2002olfactory,
  title={The olfactory receptor gene superfamily of the mouse},
  author={Zhang, Xinmin and Firestein, Stuart},
  journal={Nature neuroscience},
  volume={5},
  number={2},
  pages={124--133},
  year={2002},
  publisher={Nature Publishing Group}
}

@article{mombaerts1996visualizing,
  title={Visualizing an olfactory sensory map},
  author={Mombaerts, Peter and Wang, Fan and Dulac, Catherine and Chao, Steven K and Nemes, Adriana and Mendelsohn, Monica and Edmondson, James and Axel, Richard},
  journal={Cell},
  volume={87},
  number={4},
  pages={675--686},
  year={1996},
  publisher={Elsevier}
}

@article{ressler1993zonal,
  title={A zonal organization of odorant receptor gene expression in the olfactory epithelium},
  author={Ressler, Kerry J and Sullivan, Susan L and Buck, Linda B},
  journal={Cell},
  volume={73},
  number={3},
  pages={597--609},
  year={1993},
  publisher={Elsevier}
}

@article{ressler1994information,
  title={Information coding in the olfactory system: evidence for a stereotyped and highly organized epitope map in the olfactory bulb},
  author={Ressler, Kerry J and Sullivan, Susan L and Buck, Linda B},
  journal={Cell},
  volume={79},
  number={7},
  pages={1245--1255},
  year={1994},
  publisher={Elsevier}
}

@article{vassar1994topographic,
  title={Topographic organization of sensory projections to the olfactory bulb},
  author={Vassar, Robert and Chao, Steven K and Sitcheran, Raquel and Nu{\~n}ez, Jennifer M and Vosshall, Leslie B and Axel, Richard},
  journal={Cell},
  volume={79},
  number={6},
  pages={981--991},
  year={1994},
  publisher={Elsevier}
}

@article{jefferis2007comprehensive,
  title={Comprehensive maps of {Drosophila} higher olfactory centers: spatially segregated fruit and pheromone representation},
  author={Jefferis, Gregory SXE and Potter, Christopher J and Chan, Alexander M and Marin, Elizabeth C and Rohlfing, Torsten and Maurer Jr, Calvin R and Luo, Liqun},
  journal={Cell},
  volume={128},
  number={6},
  pages={1187--1203},
  year={2007},
  publisher={Elsevier}
}

@article{marin2002representation,
  title={Representation of the glomerular olfactory map in the {Drosophila} brain},
  author={Marin, Elizabeth C and Jefferis, Gregory SXE and Komiyama, Takaki and Zhu, Haining and Luo, Liqun},
  journal={Cell},
  volume={109},
  number={2},
  pages={243--255},
  year={2002},
  publisher={Elsevier}
}

@article{wong2002spatial,
  title={Spatial representation of the glomerular map in the {Drosophila} protocerebrum},
  author={Wong, Allan M and Wang, Jing W and Axel, Richard},
  journal={Cell},
  volume={109},
  number={2},
  pages={229--241},
  year={2002},
  publisher={Elsevier}
}

@article{price1970mitral,
  title={The mitral and short axon cells of the olfactory bulb},
  author={Price, JL and Powell, TPS},
  journal={Journal of cell science},
  volume={7},
  number={3},
  pages={631--651},
  year={1970},
  publisher={The Company of Biologists Ltd}
}

@article{deBelle1994associative,
  title={Associative odor learning in {Drosophila} abolished by chemical ablation of mushroom bodies},
  author={de Belle, J Steven and Heisenberg, Martin},
  journal={Science},
  volume={263},
  number={5147},
  pages={692--695},
  year={1994},
  publisher={American Association for the Advancement of Science}
}

@article{dubnau2001disruption,
  title={Disruption of neurotransmission in {Drosophila} mushroom body blocks retrieval but not acquisition of memory},
  author={Dubnau, Josh and Grady, Leo and Kitamoto, Toshihiro and Tully, Tim},
  journal={Nature},
  volume={411},
  number={6836},
  pages={476--480},
  year={2001},
  publisher={Nature Publishing Group}
}

@article{heisenberg1985drosophila,
  title={{Drosophila} mushroom body mutants are deficient in olfactory learning},
  author={Heisenberg, Martin and Borst, Alexander and Wagner, Sibylle and Byers, David},
  journal={Journal of neurogenetics},
  volume={2},
  number={1},
  pages={1--30},
  year={1985},
  publisher={Taylor \& Francis}
}

@article{mcguire2001role,
  title={The role of {Drosophila} mushroom body signaling in olfactory memory},
  author={McGuire, Sean E and Le, Phi T and Davis, Ronald L},
  journal={Science},
  volume={293},
  number={5533},
  pages={1330--1333},
  year={2001},
  publisher={American Association for the Advancement of Science}
}

@article{davison2011neural,
  title={Neural circuit mechanisms for pattern detection and feature combination in olfactory cortex},
  author={Davison, Ian G and Ehlers, Michael D},
  journal={Neuron},
  volume={70},
  number={1},
  pages={82--94},
  year={2011},
  publisher={Elsevier}
}

@article{miyamichi2011cortical,
  title={Cortical representations of olfactory input by trans-synaptic tracing},
  author={Miyamichi, Kazunari and Amat, Fernando and Moussavi, Farshad and Wang, Chen and Wickersham, Ian and Wall, Nicholas R and Taniguchi, Hiroki and Tasic, Bosiljka and Huang, Z Josh and He, Zhigang and others},
  journal={Nature},
  volume={472},
  number={7342},
  pages={191--196},
  year={2011},
  publisher={Nature Publishing Group}
}

@article{caron2013random,
  title={Random convergence of olfactory inputs in the {Drosophila} mushroom body},
  author={Caron, Sophie JC and Ruta, Vanessa and Abbott, LF and Axel, Richard},
  journal={Nature},
  volume={497},
  number={7447},
  pages={113--117},
  year={2013},
  publisher={Nature Publishing Group}
}

@article{li2020connectome,
  title={The connectome of the adult {Drosophila} mushroom body provides insights into function},
  author={Li, Feng and Lindsey, Jack W and Marin, Elizabeth C and Otto, Nils and Dreher, Marisa and Dempsey, Georgia and Stark, Iaroslav and Bates, Alexander Shakeel and Pleijzier, Markus William and Schlegel, Philipp and others},
  journal={eLife},
  volume={9},
  pages={e62576},
  year={2020},
  publisher={eLife Sciences Publications Limited}
}

@article{zheng2018complete,
  title={A complete electron microscopy volume of the brain of adult {Drosophila melanogaster}},
  author={Zheng, Zhihao and Lauritzen, J Scott and Perlman, Eric and Robinson, Camillia G and Nichols, Matthew and Milkie, Daniel and Torrens, Omar and Price, John and Fisher, Corey B and Sharifi, Nadiya and others},
  journal={Cell},
  volume={174},
  number={3},
  pages={730--743},
  year={2018},
  publisher={Elsevier}
}

@article{wang2021evolving,
  title={Evolving the olfactory system with machine learning},
  author={Wang, Peter Y and Sun, Yi and Axel, Richard and Abbott, LF and Yang, Guangyu Robert},
  journal={Neuron},
  volume={109},
  number={23},
  pages={3879--3892},
  year={2021},
  publisher={Elsevier}
}

@article{babadi2014sparseness,
  title={Sparseness and expansion in sensory representations},
  author={Babadi, Baktash and Sompolinsky, Haim},
  journal={Neuron},
  volume={83},
  number={5},
  pages={1213--1226},
  year={2014},
  publisher={Elsevier}
}

@article{aso2014neuronal,
  author = {Aso, Yoshinori and Hattori, Daisuke and Yu, Y. and Johnston, R. M. and Iyer, N. A. and Ngo, T. T. B. and Dionne, H. and Abbott, L. F. and Axel, R. and Tanimoto, H. and Rubin, G. M.},
  title = {The neuronal architecture of the mushroom body provides a logic for associative learning},
  journal = {eLife},
  volume = {3},
  pages = {e04577},
  year = {2014}
}

@inproceedings{Nair2010RectifiedLU,
  title={Rectified Linear Units Improve Restricted Boltzmann Machines},
  author={Vinod Nair and Geoffrey E. Hinton},
  booktitle={International Conference on Machine Learning (ICML)},
  pages={807--814},
  year={2010}
}

@inproceedings{Kingma2015AdamAM,
  title={Adam: A Method for Stochastic Optimization},
  author={Diederik P. Kingma and Jimmy Ba},
  booktitle={International Conference on Learning Representations (ICLR)},
  year={2015}
}
\bibliographystyle{iclr2025_conference}

\clearpage
\appendix

\section*{Appendix}

\section{Telugu Visualizations and Activation Dynamics}
\label{sec:telugu_appendix}

\begin{figure*}[htbp]
\centering
\begin{subfigure}[b]{0.48\textwidth}
    \centering
    \includegraphics[width=0.9\linewidth]{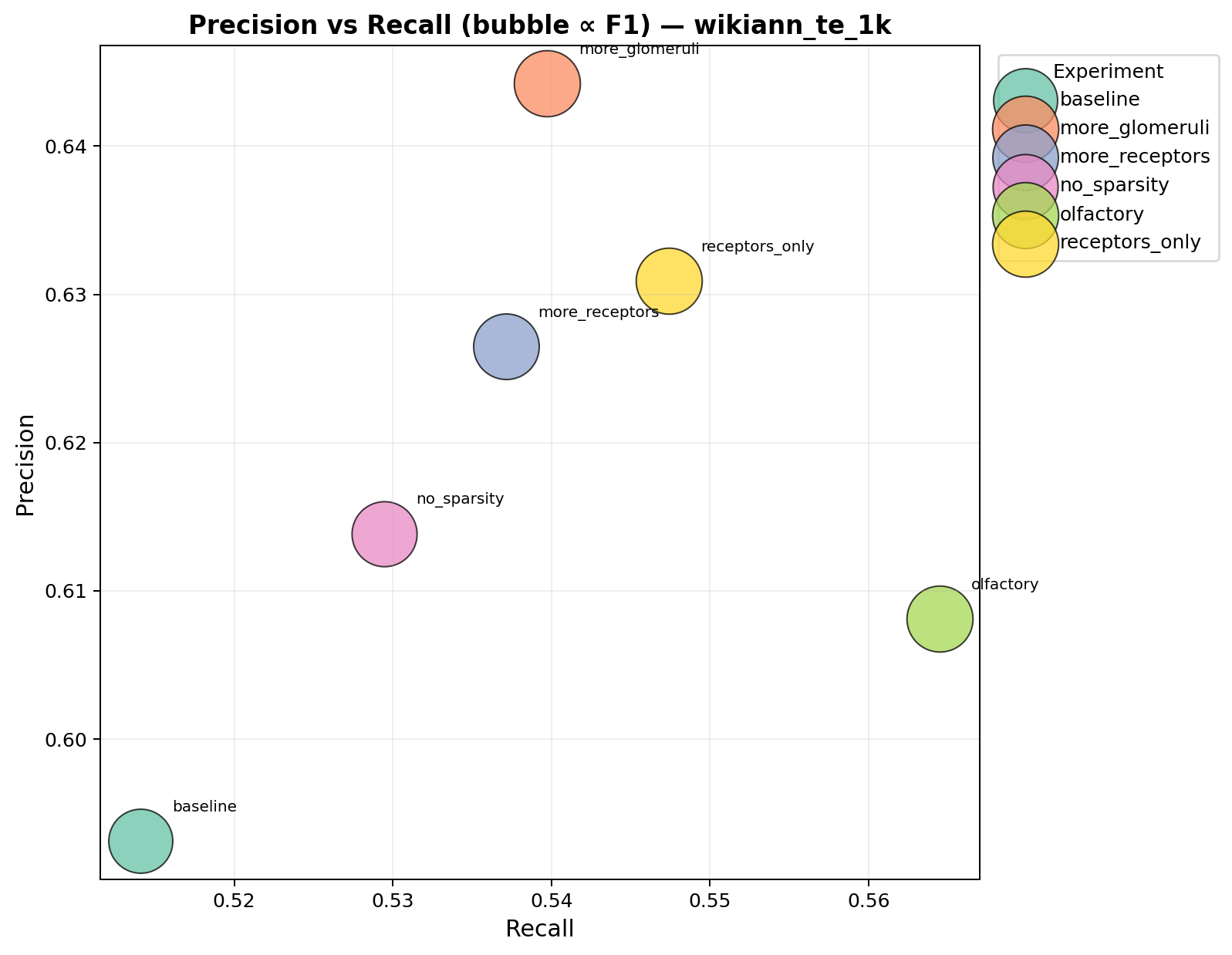}
    \caption{Precision vs. Recall bubble chart.}
    \label{fig:telugu_pr_bubble}
\end{subfigure}
\hfill
\begin{subfigure}[b]{0.48\textwidth}
    \centering
    \includegraphics[width=0.9\linewidth]{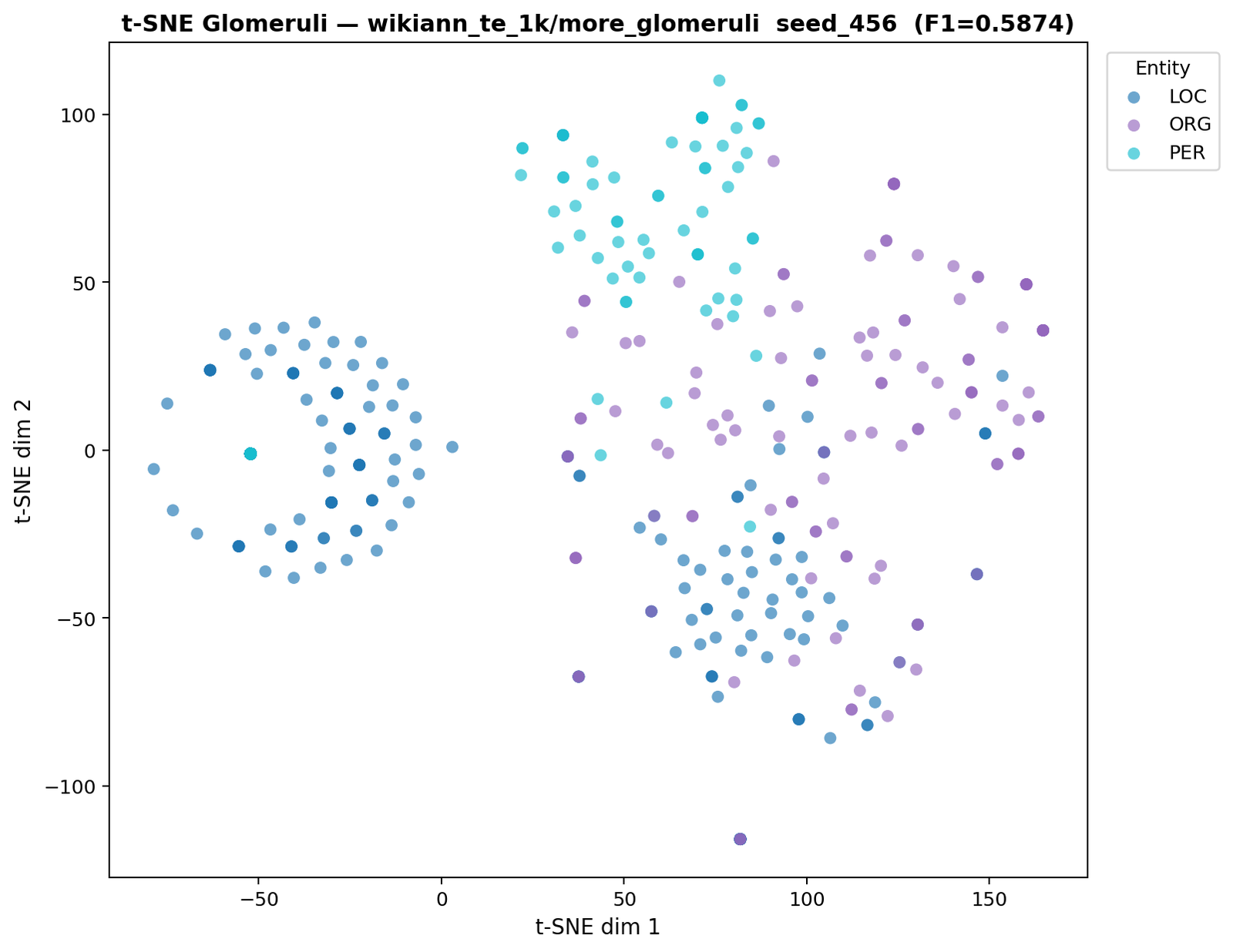}
    \caption{t-SNE visualization of token-level activations.}
    \label{fig:telugu_tsne}
\end{subfigure}
\caption{Visualizations for Telugu (1k Capped setting). (a) Precision-Recall dynamics showing model configurations. (b) t-SNE projection of the 64-dimensional glomeruli activations, showing semantic separation of named entities.}
\label{fig:telugu_visualizations_combined}
\end{figure*}

To illustrate the representational properties of our olfactory-inspired architecture in low-resource contexts, we present a consolidated visualization grid for the Telugu language under the 1k-capped training setting in Figure~\ref{fig:telugu_visualizations_combined}.

\paragraph{Figure~\ref{fig:telugu_visualizations_combined} Explanation:} 
\begin{itemize}
    \item \textbf{Precision-Recall Dynamics (a):} Plots Precision against Recall for Telugu under 1k capped training. It demonstrates that the olfactory configurations successfully shift the network into a higher-precision and higher-recall equilibrium, mitigating the typical low-precision dropoff associated with sequence models trained on very small datasets.
    \item \textbf{Semantic Clustering (b):} The emergence of clean, well-separated semantic clusters for PER, LOC, and ORG in the t-SNE 2D projection of the 64-dimensional glomeruli activations demonstrates that the representation space is highly organized and linearly separable, allowing the downstream CRF decoder to make more accurate sequence labeling decisions.
\end{itemize}

\section{Cross-Dataset F1 Performance Heatmaps}
\label{sec:cross_dataset_appendix}

\begin{figure}[htbp]
    \centering
    \includegraphics[width=0.8\linewidth]{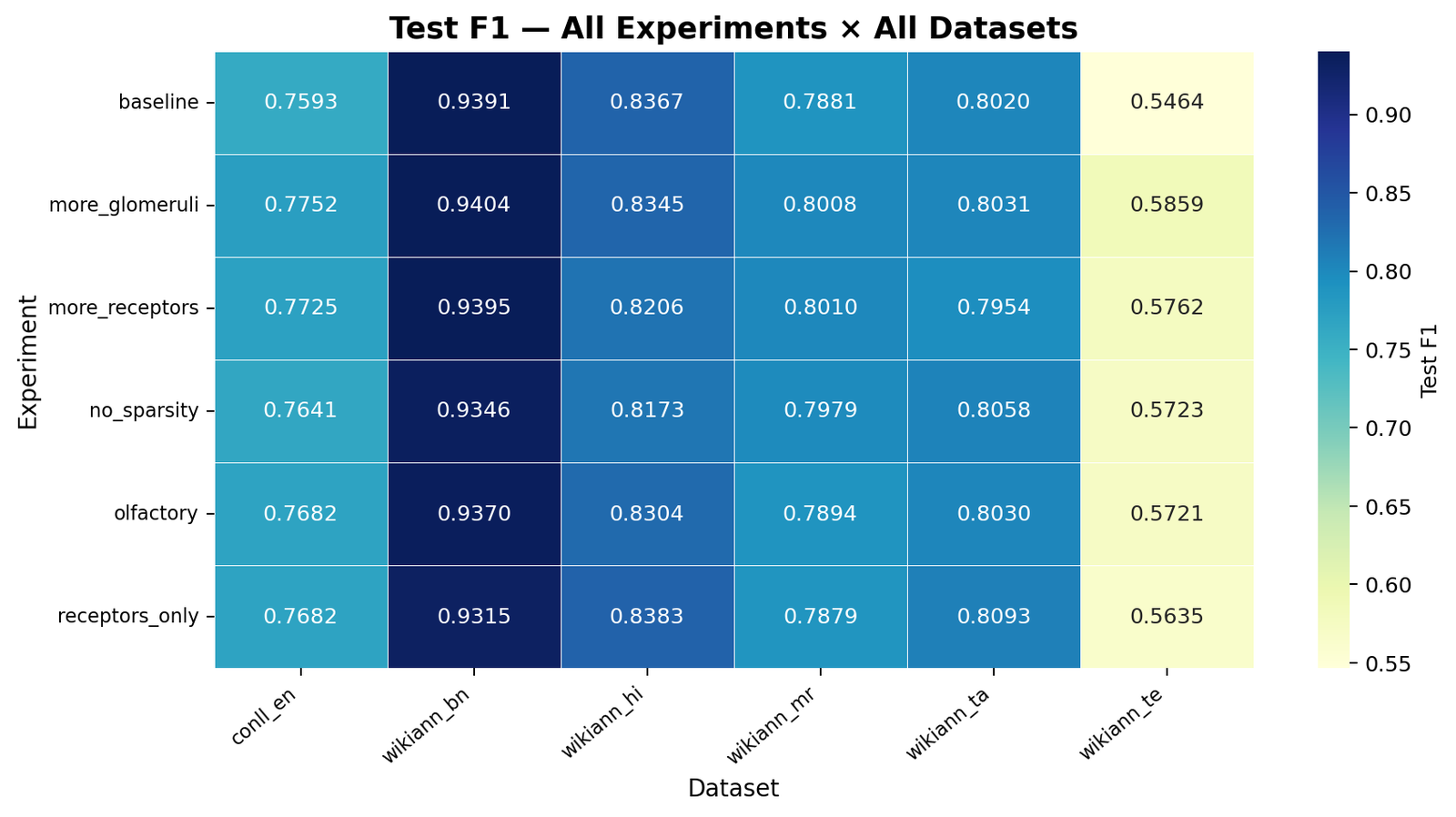}
    \caption{Cross-dataset F1 heatmap comparing all configurations (Full experiments).}
    \label{fig:cross_dataset_f1_heatmap_appendix}
\end{figure}

\begin{figure}[htbp]
    \centering
    \includegraphics[width=0.8\linewidth]{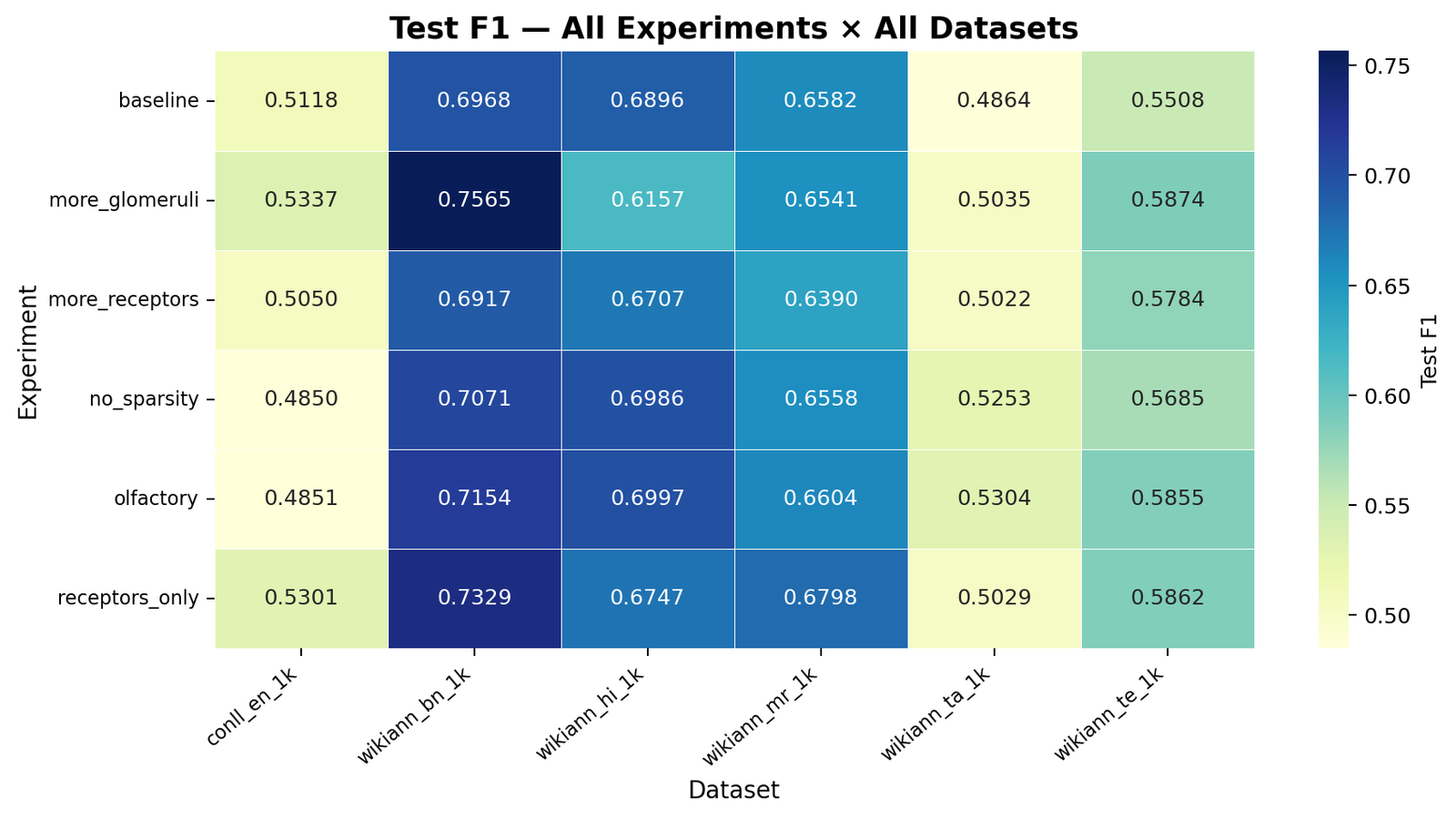}
    \caption{Cross-dataset F1 heatmap comparing all configurations (1k Capped low-resource simulation).}
    \label{fig:cross_dataset_f1_heatmap_1k_appendix}
\end{figure}

We present the cross-dataset F1 heatmaps comparing all configurations across all six datasets. Figure~\ref{fig:cross_dataset_f1_heatmap_appendix} illustrates performance under the full-scale experiments, and Figure~\ref{fig:cross_dataset_f1_heatmap_1k_appendix} illustrates performance under the 1k-capped low-resource simulation constraints.

\end{document}